\documentclass[11pt, a4paper, logo, copyright]{googlecloud}

\pdftrailerid{redacted}

\makeatletter
\renewcommand\bibentry[1]{\nocitep{#1}{\frenchspacing\@nameuse{BR@r@#1\@extra@b@citeb}}}
\makeatother
\definecolor{kmycolor}{rgb}{0.858, 0.188, 0.478}

\definecolor{llcolor}{rgb}{0.0, 0.5, 0.5}

\definecolor{yscolor}{rgb}{1.0, 0.6, 0.0}

\definecolor{yiwencolor}{rgb}{1.0, 0.0, 1.0}

\definecolor{dxlcolor}{rgb}{0.0, 0.0, 0.0}

\usepackage{kantlipsum, lipsum}
\usepackage{dsfont}

\definecolor{ourred}{HTML}{F19C99}
\definecolor{ourblue}{HTML}{7EA6E0}
\definecolor{plotred}{HTML}{f77189}
\definecolor{plotgreen}{HTML}{33b07a}
\definecolor{plotpurple}{HTML}{cc7af4}
\definecolor{plotmagenta}{HTML}{f565cc}
\definecolor{plotazure}{HTML}{38a9c5}

\definecolor{codegreen}{rgb}{0,0.6,0}
\definecolor{codegray}{rgb}{0.5,0.5,0.5}
\definecolor{codepurple}{rgb}{0.58,0,0.82}
\definecolor{backcolour}{rgb}{0.95,0.95,0.92}

\definecolor{Gray}{gray}{0.92}

\definecolor{stage1}{HTML}{34A853}
\definecolor{stage2}{HTML}{A680B8}
\definecolor{stage3}{HTML}{009999}

\usepackage{listings}

\lstdefinestyle{mystyle}{
    backgroundcolor=\color{backcolour},   
    commentstyle=\color{codegreen},
    keywordstyle=\color{magenta},
    numberstyle=\tiny\color{codegray},
    stringstyle=\color{codepurple},
    basicstyle=\ttfamily\scriptsize,
    breakatwhitespace=false,         
    breaklines=true,                 
    captionpos=b,                    
    keepspaces=true,                 
    numbersep=5pt,                  
    showspaces=false,                
    showstringspaces=false,
    showtabs=false,                  
    tabsize=2,
    frame=none,
    aboveskip=1pt,
    belowskip=1pt,
}
\usepackage[most]{tcolorbox}
\tcbset{enhanced,
boxrule=0pt,frame hidden,
borderline west={4pt}{0pt}{green!75!black},
top=0pt,bottom=0pt,
colback=green!10!white,
sharp corners}

\usepackage{colortbl}
\usepackage{float}

\usepackage{multirow}

\usepackage[authoryear, sort&compress, round]{natbib}

\usepackage[utf8]{inputenc} %
\usepackage[T1]{fontenc}    %
\usepackage{hyperref}       %
\hypersetup{
  colorlinks,
  citecolor=blue,
  linkcolor=red,
  urlcolor=blue}
\usepackage{url}            %
\usepackage{booktabs}       %
\usepackage{amsfonts}       %
\usepackage{nicefrac}       %
\usepackage{microtype}      %
\usepackage{wrapfig} %
\usepackage{graphicx} %
\usepackage{soul} %
\usepackage{enumitem}
\usepackage{amssymb,amsmath}
\usepackage{tikz, adjustbox}
\usepackage[most]{tcolorbox}
\usepackage{subcaption}
\usepackage{multirow}
\usepackage{arydshln}
\usepackage{float}
\usepackage{algorithm}
\usepackage{algorithmic}
\usepackage[capitalize,noabbrev]{cleveref}
\usepackage{fancyvrb}
\usepackage[normalem]{ulem}

\usepackage{tcolorbox}

\usepackage{amsmath,amsfonts,bm}

\def\eqref#1{equation~\ref{#1}}

\def\1{\bm{1}}

\DeclareMathAlphabet{\mathsfit}{\encodingdefault}{\sfdefault}{m}{sl}
\SetMathAlphabet{\mathsfit}{bold}{\encodingdefault}{\sfdefault}{bx}{n}

\lstdefinestyle{mystyle}{
    backgroundcolor=\color{backcolour},   
    commentstyle=\color{codegreen},
    keywordstyle=\color{magenta},
    numberstyle=\tiny\color{codegray},
    stringstyle=\color{codepurple},
    basicstyle=\ttfamily\scriptsize,
    breakatwhitespace=false,         
    breaklines=true,                 
    captionpos=b,                    
    keepspaces=true,                 
    numbers=left,                    
    numbersep=5pt,                  
    showspaces=false,                
    showstringspaces=false,
    showtabs=false,                  
    tabsize=2,
    frame=none,
    aboveskip=1pt,
    belowskip=1pt,
}
\lstdefinestyle{plainins}{
    backgroundcolor=\color{white},   
    commentstyle=\color{codegreen},
    keywordstyle=\color{magenta},
    numberstyle=\tiny\color{codegray},
    stringstyle=\color{codepurple},
    basicstyle=\ttfamily\scriptsize,
    breakatwhitespace=false,         
    breaklines=true,                 
    captionpos=b,                    
    keepspaces=true,                 
    numbers=none,                    
    numbersep=5pt,                  
    showspaces=false,                
    showstringspaces=false,
    showtabs=false,                  
    tabsize=2,
    aboveskip=0pt,
    belowskip=0pt,
    frame=single
}
\lstdefinestyle{plainexam}{
    backgroundcolor=\color[HTML]{FFFCF3},   
    commentstyle=\color{codegreen},
    keywordstyle=\color{magenta},
    numberstyle=\tiny\color{codegray},
    stringstyle=\color{codepurple},
    basicstyle=\ttfamily\scriptsize,
    breakatwhitespace=false,         
    breaklines=true,                 
    captionpos=b,                    
    keepspaces=true,                 
    numbers=none,                    
    numbersep=5pt,                  
    showspaces=false,                
    showstringspaces=false,
    showtabs=false,                  
    tabsize=2,
    aboveskip=0pt,
    belowskip=0pt
}

\newcommand{\model}{\textsc{FVScore}}
\newcommand{\dataset}{\textsc{FVBench}}

 \newcounter{finding}

\title{\model{}: Specification-Driven Neuro-Symbolic Video Evaluation}

\correspondingauthor{Do Xuan Long: xuanlong.do@u.nus.edu; Long T. Le, Yale Song: \{longtle, yalesong\}@google.com}

\renewcommand{\today}{}

\author[1 2 *]{Do Xuan Long}
\author[1]{Long T. Le}
\author[1]{Yiwen Song}
\author[2]{Min-Yen Kan}
\author[1]{Tomas Pfister}
\author[1]{Yale Song}
\affil[1]{Google Cloud AI Research}
\affil[2]{National University of Singapore}

\newcommand{\copyrightext}{\footerfont * This work was done while Do Xuan Long was a Student Researcher at Google Cloud AI Research. Our project is available at \href{https://dxlong2000.github.io/FVSCORE/}{this url.}}

\renewcommand{\today}{}

\usepackage[table]{xcolor}

\usepackage{wrapfig}
\usepackage{amsfonts}       %
\usepackage{nicefrac}       %
\usepackage{microtype}      %
\usepackage{wrapfig} %
\usepackage{graphicx} %
\usepackage{soul} %
\usepackage{enumitem}
\usepackage{amssymb,amsmath}
\usepackage{tikz, adjustbox}
\usepackage[most]{tcolorbox}
\usepackage{subcaption}
\usepackage{multirow}
\usepackage{arydshln}
\usepackage{float}
\usepackage{algorithm}
\usepackage{algorithmic}

\usepackage{fancyvrb}
\usepackage{fvextra}

\fvset{
  breaklines=true,
  breaksymbol={},
  fontsize=\scriptsize,
  frame=single
}
\usepackage[normalem]{ulem}

\usepackage{booktabs}

\usepackage{tcolorbox}

\definecolor{kmycolor}{rgb}{0.858, 0.188, 0.478}
\definecolor{llcolor}{rgb}{0.0, 0.5, 0.5}

\definecolor{yscolor}{rgb}{1.0, 0.6, 0.0}
\definecolor{yiwencolor}{rgb}{1.0, 0.0, 1.0}

\definecolor{dxlcolor}{rgb}{0.0, 0.0, 0.0}

\usepackage{url}

\usepackage[capitalize,noabbrev]{cleveref}

\begin{document}
\begin{abstract}
We introduce \textbf{Faithfulness Video Score (\model{})}, a neuro-symbolic framework for specification-driven video generation evaluation. Existing evaluators rely on holistic MLLM judgments or vision--language metrics, yet aligning their scores with human judgments and tracing them to specific temporal violations remain open challenges. \model{} addresses these limitations by compiling each video into explicit propositions capturing prompt requirements and implicit propositions capturing unstated physical, temporal, and cross-modal constraints. It grounds each proposition using specialized perceptual tools and evaluates the resulting confidence traces with soft temporal operators inspired by Linear Temporal Logic. A Perception-Aware Calibration step then aligns verification with human perception by retaining only perceptually significant violations before producing a timestamp- and evidence-grounded report. We conduct extensive experiments across public and complex multi-event benchmarks with the strongest video models to date, demonstrating that \model{} outperforms holistic and agent-based evaluators in human correlation while providing more accurate and interpretable feedback.
\end{abstract}

\maketitle

\section{Introduction}

\begin{wrapfigure}{r}{0.47\textwidth}
    \centering
    \vspace{-5mm}
    \includegraphics[width=\linewidth]{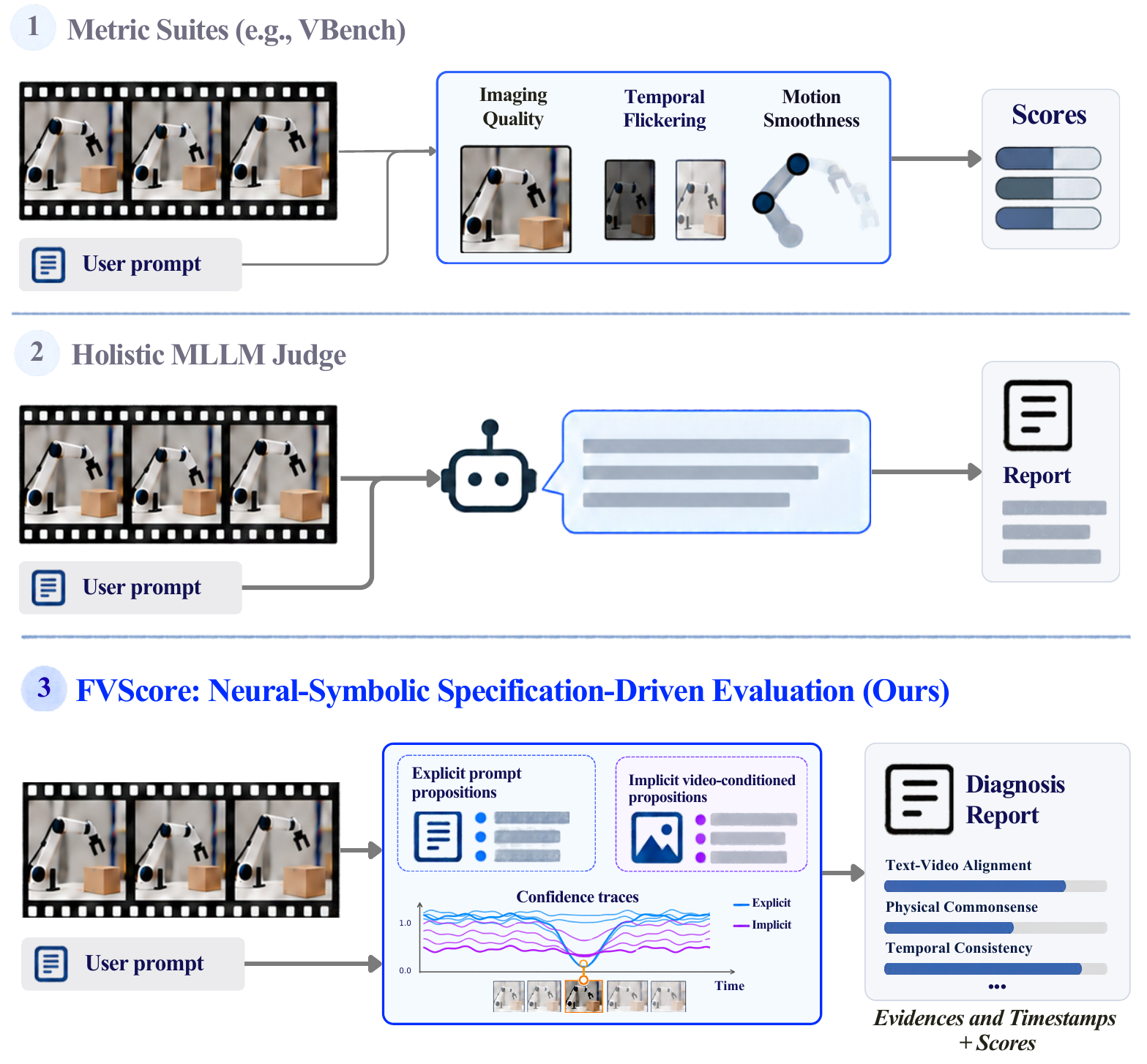}
    \caption{\textbf{Three paradigms for video evaluation.} Metric suites score predefined dimensions, while MLLM judges produce reasoning reports. \model{} compiles explicit and video-conditioned implicit propositions, evaluates their confidence traces, and reports dimension scores with timestamped evidence.}
    \label{fig:teaser}
\end{wrapfigure}


Automatic video evaluation is a central yet unresolved problem in generative AI. As text-to-video (T2V) models~\citep{deepmind2025veo3,seedance2} attain increasingly outstanding visual realism, judging a generation primarily by visual quality is no longer sufficient. A generated video can look sharp and cinematic, yet appear nonsensical to human viewers at the same time. For example, a video may misbind a requested action (\emph{text--video alignment}), lose character identity over time (\emph{temporal consistency}), violate basic physical laws (\emph{physical commonsense}), or desynchronize audio from its visual source (\emph{video--audio alignment})~\citep{tilak2026brite,zheng2025locot2v,sharan2025neuro,guo2025t2vphysbench}. Together, these constitute the essential dimensions of \textbf{faithful generation}. Evaluating them requires determining not only explicit user requirements, but also implicit commonsense behaviors that humans expect.

Existing studies have addressed important parts of this challenge. Early metrics compare distributional or joint-embedding features~\citep{unterthiner2018towards,hessel2021clipscore} for text--video alignment, while comprehensive benchmarks such as VBench, EvalCrafter, and T2VScore evaluate videos using criteria-based proxies~\citep{Huang_2024_CVPR,liu2024evalcrafter,sun2024t2v}. Because such metrics can align weakly with human ratings and offer limited interpretability, recent approaches train evaluators directly on human scores~\citep{he-etal-2024-videoscore,he2025videoscore2}, use MLLM-based multi-agent critiques for better interpretability~\citep{chen2024mllm,long2026vista,long20262}, or decompose prompts into atomic visual checks~\citep{hu2023tifa,cho2024davidsonian,sharan2025neuro,song2026vqqa} (\Cref{fig:teaser}). Nevertheless, they remain limited by predefined evaluation criteria or user input requirements. To our knowledge, no evaluator explicitly derives \emph{which} implicit temporal constraints emerge from video events, nor accurately verifies \emph{when} they are violated. Consequently, perceptually salient violations, such as implausible state transitions or video--audio mismatches, are often missed or folded into holistic scores, yielding diagnoses too general to identify what failed and when.

To address these limitations, we formulate video evaluation as \textbf{Specification-Driven Verification}. Inspired by formal verification~\citep{pnueli1977temporal,baier2008principles}, we use specification languages to evaluate videos on specific faithfulness constraints, enabling targeted failure localization. We introduce \textbf{\model{}}, a neuro-symbolic framework that separates this process into \emph{compilation} and \emph{execution}. The compiler generates \emph{explicit propositions} from the user prompt and \emph{implicit propositions} from observed video dynamics. During execution, \model{} routes these to specialized neural or symbolic verifiers operating over a temporally indexed video representation. It then evaluates their confidence traces using soft LTL-inspired operators to identify {what} fails and {when}. Finally, a Perception-Aware Calibration step retains perceptually significant violations to align scores with humans and produce timestamp-grounded diagnoses. Our comprehensive evaluations show that \model{} significantly outperforms existing evaluators, increasing human correlation by up to $17\%$ on alignment, $26\%$ on physical commonsense, $20\%$ in consistency, and $44\%$ on video--audio sync, alongside $55\%$ and $50\%$ average win margins in groundedness and completeness of the feedback. In summary, our contributions are:

\begin{itemize}
    \item We formulate video evaluation as \textbf{Specification-Driven Verification}, deriving both explicit (prompt-stated) and implicit (video-conditioned) requirements for faithfulness video evaluation.
    \item We introduce \textbf{\model{}}, an automatic framework that compiles these requirements into propositions, routes them to specialized verifiers, evaluates their confidence traces, and calibrates scores for human alignment.
    \item \model{} achieves significantly stronger human alignment than prior evaluators (e.g., 26\% and 44\% correlation gains on physical commonsense and video--audio sync) while providing more grounded and useful diagnoses.
\end{itemize}

\section{Related Work}

\paragraph{Generative Video Evaluation.}  
Automatic video evaluation has achieved remarkable progress, developed along four lines. Earlier efforts develop distribution and similarity metrics~\citep{salimans2016improved,heusel2017gans,hessel2021clipscore}, but reduce video quality or text--video alignment to a single scalar. Extending them, benchmark suites cover comprehensive dimensions~\citep{liu2023fetv,Huang_2024_CVPR,huang2025vbench,liu2024evalcrafter,sun2024t2v,hua2026vabench,zhou2026avgen}, yet aligning their scores to human judgments remains a significant challenge~\citep{sharan2025neuro}. To improve it, recent work trains specialized evaluators directly on human ratings~\citep{he-etal-2024-videoscore,bansal2025videophy,he2025videoscore2,bansal2026videophy,liu2026improving} or employs MLLMs as judges~\citep{zheng2023judging,chen2024mllm,long2026vista,long20262}; nevertheless, their evaluation remains unstable across models, scenarios, and rubric choices. Recent methods decompose the prompt into atomic verifications~\citep{hu2023tifa,cho2024davidsonian,guan2025etva,song2026vqqa}, leading to more interpretable evaluation. 
Closest to us, NeuS-V~\citep{sharan2025neuro} verifies text--video alignment through formal verification. However, we find that it correlates weakly with human judgments on public benchmarks, as it can compound joint probabilities over frames (\Cref{sec:main-experiments}), remains restricted to a single criterion, and produces no grounded explanation. \model{} instead provides a unified interface that verifies both explicit and implicit propositions over a temporal video representation. This allows it to identify what fails and when, providing more interpretable and aligned evaluation while can be readily extended to future evaluation criteria.

\paragraph{Neuro-Symbolic Reasoning.} 
Neuro-symbolic reasoning bridges the gap between neural network's pattern recognition and formal logic's interpretability \citep{hu2016harnessing,yi2018neural,yu2023survey,lu2024surveying,mao2019neuro}. These methods separate perception from reasoning: a neural front-end turns visual evidence into a structured representation over which a symbolic back-end reasons, whether through learned program execution~\citep{yi2018neural,mao2019neuro} or training-free program synthesis~\citep{gupta2023visual,suris2023vipergpt}. This paradigm has been extended to video through spatio-temporal causal reasoning benchmarks~\citep{yi2020clevrer}, and recently to text--video alignment by NeuS-V~\citep{sharan2025neuro}, which compiles the prompt into a temporal-logic specification and verifies it against a probabilistic video automaton. However, its automaton grows exponentially with propositions, and calculating satisfaction as a joint probability over-penalizes videos as more frames are evaluated. Furthermore, NeuS-V derives logical programs from prompt only. In contrast, {\model{} dynamically generates implicit propositions} from the video itself. By verifying each proposition independently against its own confidence trace, \model{} {prevents a single failure from artificially depressing the overall score.}


\section{Methodology}

\subsection{Problem Formulation} \label{sec:formulations}

Given a video $V$ synthesized from a diffusion model using a natural-language prompt $P$, our goal is to evaluate its \textbf{faithfulness}. Following recent literature, we define faithfulness across four essential dimensions: \textbf{text--video alignment} \citep{Huang_2024_CVPR, liu2024evalcrafter, sun2024t2v}, \textbf{physical commonsense} \citep{bansal2025videophy,guo2025t2vphysbench}, \textbf{temporal consistency} \citep{zheng2025locot2v, tilak2026brite}, and \textbf{video--audio synchronization} \citep{hua2026vabench, zhou2026avgen}. In addition, it is worth noting that while text--video alignment is evaluated against requirements explicitly stated in prompt $P$, the remaining dimensions rely on implicit physical and causal constraints dictated by the specific events generated in $V$. An ideal evaluator therefore must dynamically derive both these explicit and implicit requirements and judge them in alignment with humans.

\begin{figure*}[t!]
\centering
\includegraphics[width=\textwidth]{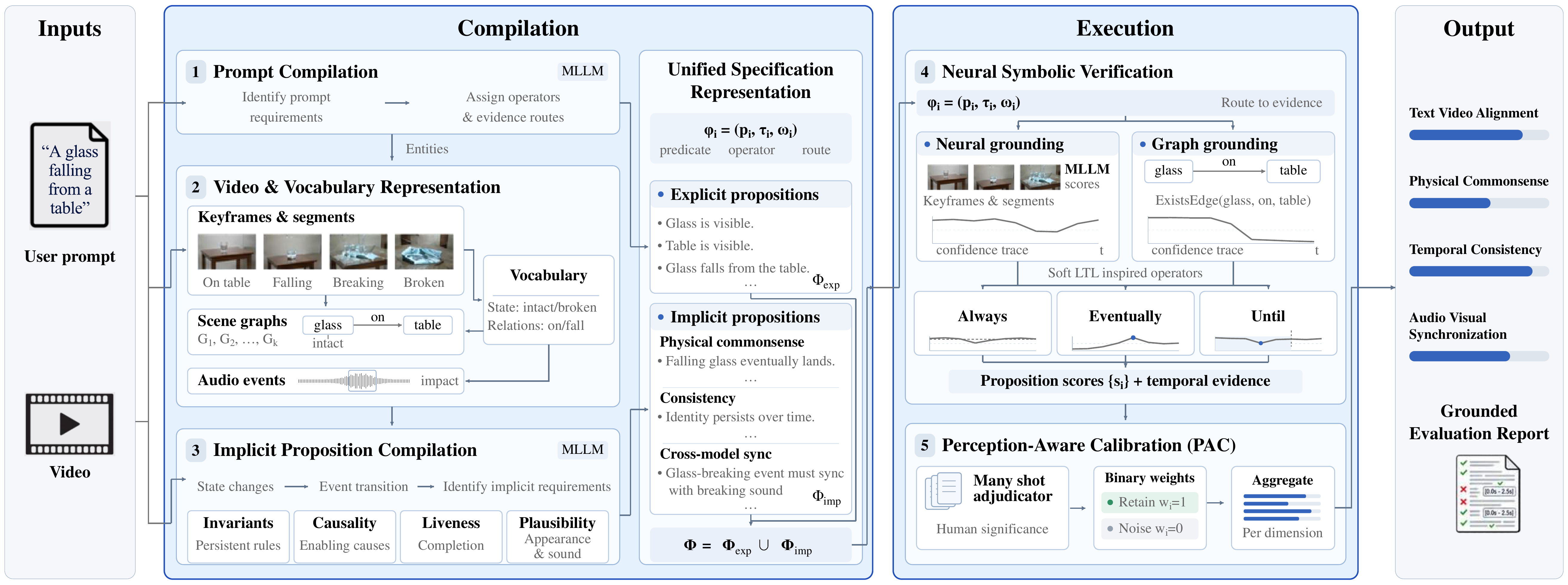}
\vspace{-3mm}
\caption{\textbf{\model{}}, which compiles the user prompt and video observations into explicit and implicit propositions, verifies them via specialized neural and symbolic tools, and applies Perception-Aware Calibration to produce timestamp-grounded scores and reports.}
\label{fig:logicvideoeval-overview}
\end{figure*}

\paragraph{Specification-Driven Verification.}
We formulate video evaluation as {Specification-Driven Verification}, in which the evaluator first compiles a set of situation-specific specifications $\Phi$ and then verifies them against perceptual evidence and generates the report. Formally,
\vspace{1mm}
\begin{equation}\label{eq:problem-formulation}
\begin{aligned}
&\Phi := \operatorname{Compile}(P, V),\\
&S(V,P) := \operatorname{Execute}(\Phi, V)
\in [0,1]^N \times {E}
\end{aligned}
\end{equation}
\vspace{1mm}
Here, $S$ denotes the evaluator, which returns a score in $[0,1]$ for each of the $N$ dimensions and an explanation $E^{d} \in {E}$ grounded in specific propositions and timestamps. Unlike holistic evaluators, $\Phi$ is not predefined; instead, it is situation-dependent and derived from the prompt and video.

\subsection{The \model{} Framework}

\Cref{fig:logicvideoeval-overview} outlines our proposed \model{} which leverages a Multimodal Large Language Model (MLLM) as the core engine. It realizes $S$ through a \emph{Compilation} phase that produces the specification $\Phi$ and an \emph{Execution} phase that verifies it and generates the report. Compilation occurs in two steps: Step 1 derives explicit propositions from $P$, while Step 3 derives implicit propositions from a representation $\mathcal{R}$ constructed in Step 2. Execution then verifies each proposition to obtain its score (Step 4), then identifies perceptually significant violations, aggregates dimension scores, and produces a grounded report (Step 5). We now formalize $\Phi$ and the unified temporal language used to express it.

\paragraph{Component Specifications.}
To evaluate faithfulness, we structure $\Phi$ into explicit and implicit propositions:
\begin{equation}
  \Phi := \Phi_{\text{exp}} \cup \Phi_{\text{imp}}
  \label{eq:phi}
\end{equation}

Here, $\Phi_{\text{exp}}$ contains propositions for the entities, events, orderings, and timings that $P$ explicitly requires; $\Phi_{\text{imp}} := \Phi^{\text{phy}}_{\text{imp}} \cup \Phi^{\text{cons}}_{\text{imp}} \cup \Phi^{\text{sync}}_{\text{imp}}$ captures the \emph{(i) physical commonsense} ($\Phi^{\text{phy}}_{\text{imp}}$) implied by the depicted situation but unstated in $P$, covering both visual and auditory dynamics, \emph{(ii) consistency} ($\Phi^{\text{cons}}_{\text{imp}}$), which enforces the coherence of the video's entities and audio, and \emph{(iii) synchronization} ($\Phi^{\text{sync}}_{\text{imp}}$), which enforces temporal and semantic correspondence between detected sounds and concurrent visual events. Each proposition records the modality of evidence it needs: visual, auditory, or cross-modal. Each dimension is then computed from its corresponding propositions: text--video alignment from $\Phi_{\text{exp}}$, physical commonsense from the visual ones of $\Phi^{\text{phy}}_{\text{imp}}$, consistency from $\Phi^{\text{cons}}_{\text{imp}}$, and synchronization from $\Phi^{\text{sync}}_{\text{imp}}$.

\paragraph{A Unified Specification Representation.}
We design a unified representation where every requirement from all dimensions is compiled into a proposition $\phi_i=(p_i, \tau_i, \omega_i)$. This encapsulates an atomic predicate $p_i$, a Linear Temporal Logic (LTL) operator $\tau_i \in \{\textsc{\small{Always}}, \textsc{\small{Eventually}}, \textsc{\small{Until}}\}$, and a verification route $\omega_i$. The route dispatches the predicate to the appropriate grounding track (either neural or graph; detailed in Step 4) and verification tool, which yields a confidence trace over the video. The temporal operator then reduces this trace into a single proposition score. By standardizing evaluation into this interface, \model{} can easily extend to new dimensions simply by adding new tools. We detail the specific parameterizations in \Cref{app:unified_language}. 

\subsubsection{Compilation}

\paragraph{Step 1. Prompt Compilation for $\Phi_{\text{exp}}$ (Text-Video Alignment).}\label{sec:prompt_compilation}   
This step converts the prompt into explicit propositions and assigns each predicate a temporal operator and an evidence route from the tool space $\Omega$. The tool space includes visual, audio, and MLLM tools specializing in object detection, depth, shot scale, and diarization, see \Cref{sec:graph_construction}. Given $P$, $\Omega$, and the supported temporal operators, the MLLM-based compiler produces:
\begin{equation}
  \Phi_{\text{exp}},\; \mathcal{E}
  \leftarrow \operatorname{Compile}^{\text{exp}}_{\mathrm{MLLM}}(P;\Omega)
\end{equation}
The predicates are compiled only when they are stated explicitly in $P$. We also return an auxiliary entity set $\mathcal{E}$ containing the concrete entities referenced by the propositions, which is used to ground the video representation in subsequent steps.

\paragraph{Step 2. Video and Vocabulary Representation.}
This step grounds the visual and auditory content of $V$ into a structured representation $\mathcal{R}$ that exposes keyframes, video segments, entity states, relations, and sound events on a video timeline. 

\textbf{Keyframe Layer.} 
We sample keyframes $I_{1:K}$ from $V$ and retain their frame indices. Each keyframe is paired with the video segment extending from the preceding keyframe to the current one, with the first segment beginning at the start of the video. The keyframes provide evidence for appearance and entity state, while the corresponding segments provide evidence for object and camera motions.

\textbf{Entity-State Vocabulary.} Using entity set $\mathcal{E}$ and the sampled keyframes $I_{1:K}$, an MLLM constructs a vocabulary $\mathcal{V}$ for the video world: 
\begin{equation}
  \mathcal{V}
  \leftarrow
  \operatorname{Vocab}_{\mathrm{MLLM}}
  (P,\mathcal{E},I_{1:K})
\end{equation}
For each entity, $\mathcal{V}$ specifies observable state dimensions and their admissible values (e.g., $\textit{open\_state}\in\{\textit{open},\textit{closed}\}$ for a door). It also specifies admissible action relations as base-form verbs, such as \textit{open} and \textit{hold}. Constraining all keyframes to $\mathcal{V}$ gives entity states and action relations a common set of symbols, allowing propositions to be subsequently verified symbolically.

\textbf{Scene-Graph Layer.} 
To verify temporal visual states within the video, we construct an ordered sequence of per-keyframe scene graphs $\mathcal{G}=\{G_k\}_{k=1}^{K}$. Each $G_k$ records entity-state assignments and action relations inferred from the corresponding keyframe and constrained to $\mathcal{V}$. The graph can also be augmented with tool-specific relation and attribute evidence, such as depth ordering and shot scale. Details of graph construction and the associated perceptual tools are provided in \Cref{sec:graph_construction}.

\textbf{Audio Layer.} 
For a video with an audio stream, we extract a mono waveform and apply specialized audio and MLLM tools~\citep{park2024sortformer,googledeepmind2025gemini3flash} to identify foreground sound events, each represented by its start time, end time, description, and confidence. When $\Phi_{\mathrm{exp}}$ contains auditory propositions, their predicate texts are supplied as the target sound vocabulary. The audio layer is empty when no audio stream is available.

\paragraph{Step 3. Implicit Proposition Compilations.} 
This step expands $\Phi$ with implicit propositions $\Phi_{\text{imp}}$ that capture unstated commonsense, consistency, and cross-modal requirements.

\textbf{Physical Commonsense Propositions $\Phi^{\text{phy}}_{\text{imp}}$.} 
To extract the unstated physical constraints implied by the video's dynamics, we first convert the entity-state sequence in $\mathcal{G}$ into an event-transition narrative by retaining only the state dimensions that change across keyframes. Nevertheless, extracting the implicit physical rules that govern this narrative sequence is challenging. Because the space of possible rules is open-ended, naively prompting an MLLM tends to produce many trivial or redundant rules, overshadowing those violations that actually matter to human perception. To systematically discover those that matter most to humans, motivated by the \textbf{Temporal-Property Hierarchy Theory}~\citep{manna1990hierarchy}, \model{} generates physical propositions across four core categories: \textbf{(i) Invariants}, which enforce persistent constraints such as mutual exclusion and count stability; \textbf{(ii) Causality}, which requires an enabling cause to be present when its effect occurs; and \textbf{(iii) Liveness}, which requires a physical process that has started to reach its expected completion. We additionally generate \textbf{(iv) Plausibility} propositions for plausible entity appearance, layout, and sounds. Each proposition is strictly assigned to one category to avoid duplication, while prompt restatements and trivial rules are excluded:
\vspace{1mm}
\begin{equation}
  \Phi^{\text{phy}}_{\text{imp}}
  \leftarrow
\operatorname{Compile}^{\mathrm{imp}}_{\mathrm{MLLM}}
  (P,\mathcal{E},\mathcal{V},G, \Omega)
\end{equation}
\vspace{1mm}
\indent \textbf{Consistency Propositions  $\Phi^{\text{cons}}_{\text{imp}}$.}  
For each entity in $\mathcal{E}$, we define a template-based proposition by requiring its identity or layout to remain coherent across keyframes while permitting natural changes in pose, motion, tone, or lighting. A separate proposition enforces global background stability, explicitly prohibiting abrupt scene cuts, objects that inexplicably appear or vanish, and temporal flickering. 

\textbf{Synchronization Propositions $\Phi^{\text{sync}}_{\text{imp}}$.}
For each detected foreground sound event in $\mathcal{R}$, we define a template-based proposition requiring the sound to be temporally synchronized with the visual content during its detected interval. Each proposition retains the sound description and its start and end times, which is then evaluated with the corresponding video segment for synchronization.

\subsubsection{Execution}

\paragraph{Step 4. Neural-Symbolic Proposition Verifications.} 
This step verifies $\Phi$ to produce the  scores and reasoning.

\textbf{Neural Predicate Grounding.}
For each $\phi_i \in \Phi$ routed for neural verification, we ground its atomic predicate $p_i$ in visual evidence using either an MLLM or specialized neural models. Depending on the property, both explicit and implicit propositions are evaluated against discrete keyframes (e.g., for static entity presence and counting) or continuous video segments (e.g., for motion), producing frame-level confidence scores. Consistency yields per-entity and background scores over the ordered keyframes, while video--audio sync yields per-sound synchronization scores from the associated muted video segments. Each dimension retains its own score representation for subsequent aggregation.

\textbf{Graph Predicate Grounding.} 
For propositions routed to graph-based grounding, we compile their predicates into formal queries using the MLLM, strictly grounded on $\mathcal{V}$, and evaluate them against scene graph $\mathcal{G}$. The query language represents entity presence and subject relations using \textsc{Exists} and \textsc{ExistsEdge}, and composes multiple conditions using \textsc{And} and \textsc{Or}. Together, the neural and graph-based routes provide each proposition with a confidence trace for symbolic verification.

\textbf{Symbolic Verification.} We next evaluate the confidence traces using the temporal operators $\tau_i$. Unlike prior methods~\citep{sharan2025neuro} that compute joint probabilities over video automata, which scales exponentially with the number of propositions and whose joint-probability computation can compound uncertainty as more frames and propositions are evaluated, we verify each proposition independently using \textbf{soft LTL-inspired operators}: \textsc{Always} averages confidence over the full video or its specified temporal window, \textsc{Eventually} retains the maximum confidence, and \textsc{Until} tracks the minimum confidence that the preceding predicate holds or occurs before the consequent occurs, while checking any specified response delay or repetition requirement. We apply a temporal tolerance of $t=0.1$s for strict timing constraints like windows or deadlines, yielding a score $s_i$ for each proposition $\phi_i \in \Phi_d$ for dim $d$, which are then passed to the next step for perceptual calibration and dimension aggregation.

\paragraph{Step 5. Calibration and Report.} 
While proposition scores provide useful insights into video faithfulness, they do not explicitly account for the perceptual significance of individual violations, which limits their alignment with human judgments. We introduce \textbf{PAC}, a simple proposition-level \textbf{Perception-Aware Calibration} mechanism for alignment. We employ a many-shot in-context MLLM adjudicator to assign each proposition and its score a binary weight according to whether violating it would be \emph{perceptually significant} to humans ($w_i=1$) or \emph{negligible noise} ($w_i=0$). To construct this many-shot context, we sample violated propositions from highly rated and poorly rated videos as negligible noise and perceptually significant demonstrations respectively from a held-out set. We then cap the maximum demonstrations (20 each type). After obtaining $w_i$, defining $\Phi^{*}_d=\{\phi_i\in\Phi_d:w_i=1\}$, the final score for dim $d$ is:
\begin{equation}
\text{score}_{d} = \operatorname{Agg}_d\big(\{\, s_i : \phi_i \in \Phi^{*}_d \,\}\big),
\end{equation}
where \(\operatorname{Agg}_d\) aggregates the proposition scores relevant to each dimension, with \(\operatorname{Agg}_d(\emptyset):=1\). We implement \(\operatorname{Agg}\) as the average; PAC can either increase or decrease aggregated scores. Finally, \model{} synthesizes a natural-language Evaluation Report from the PAC-filtered propositions, summarizing what succeeded and failed and grounding every claim in specific propositions, verification rationales, and timestamps (in seconds).

\section{Experiments}\label{sec:main-experiments}

We conduct extensive experiments to answer five key research questions: \textbf{(RQ1)} \emph{Does our neuro-symbolic approach improve alignment with human judgements compared to existing evaluators?} \textbf{(RQ2)} \emph{Does proposition-level decomposition yield more accurate, grounded, and useful qualitative feedback?} \textbf{(RQ3)} \emph{How do specific symbolic verification tools and temporal logic operators contribute to the framework's overall performance?} \textbf{(RQ4)} \emph{How accurately does the compiler generate and orchestrate explicit and implicit propositions across specialized verification tools?} \textbf{(RQ5)} \emph{To what extent does PAC successfully filter imperceptible noise to align strict formal verification with human perception?}
\subsection{Experimental Setups}

\paragraph{Baselines.} We compare \model{} against seven video evaluators grouped into four widely used paradigms. 
(i)~\emph{Evaluation Suites:} \textbf{VBench}~\citep{Huang_2024_CVPR}, a standard comprehensive suite for video generation evaluation. 
(ii)~\emph{Advanced MLLM Reasoning:} Methods that enhance MLLM reasoning, including \textbf{MLLM-Judge w/ Self-Consistency}~\citep{wang2022self}; \textbf{Self-Refine}~\citep{madaan2023self}, which iteratively generates and incorporates its own critiques up to 10 iterations; and \textbf{Multi-Agent Debate}~\citep{du2024improving}, where multiple MLLM instances debate to reach a consensus score up to 10 rounds. 
(iii)~\emph{Agentic Frameworks:} \textbf{VISTA's MMAC}~\citep{long2026vista}, a jury of specialized agents scoring along visual, audio, and contextual dimensions; and \textbf{VQQA}~\citep{song2026vqqa}, a question-answering evaluator that scores via prompt-derived queries. 
(iv)~\emph{Neuro-Symbolic Methods:} \textbf{NeuS-V}~\citep{sharan2025neuro}, a temporal-logic evaluation method targeting text--video alignment. For all MLLM-based baselines, we use \textbf{Gemini~3 Flash}~\citep{googledeepmind2025gemini3flash} as the default model. We also experiment with \textbf{four other SOTA omni models} in \Cref{appdx:experiments-other-mllms} and additional baselines in \Cref{appdx:additional-baselines}.

\paragraph{Benchmarks.} 
We evaluate \model{} on two primary objectives: \textbf{score alignment} with human judgments and qualitative \textbf{feedback quality}. To ensure comprehensiveness, we test across both single-event and complex multi-event scenarios. For single-event evaluation, we sample 900 prompts with per-video human ratings from \textbf{EvalCrafter}~\citep{liu2024evalcrafter}, \textbf{VideoFeedback}~\citep{he-etal-2024-videoscore}, and \textbf{VideoFeedback2}~\citep{he2025videoscore2} (300 prompts each). For complex scenarios, we include \textbf{VBench 2.0}~\citep{zheng2025vbench}, which contains 320 multi-step prompts requiring coherent sequential ordering. However, because VBench 2.0 lacks per-video human ratings, it can only be used to evaluate feedback quality. Furthermore, it lacks the dense causality, strict response deadlines, and cross-modal constraints required to stress-test the faithfulness. To bridge these gaps, we introduce \textbf{\dataset{}}, a testing benchmark that features 200 compositional prompts with: (i) \textbf{Cross-modal multi-event instructions:} dense scenarios mapped to classical temporal properties;
and (ii) \textbf{Very high-quality videos:} 400 videos synthesized by SOTA video models \citep{seedance2,deepmind2025veo3}. Our construction details are outlined in \Cref{subsec:dataset-annotations}. 
We provide an end-to-end example of \model{} in our supplementary materials.

\paragraph{Metrics.} 
For correlations, following \citet{sharan2025neuro}, we measure \textbf{Pearson ($P$)} and \textbf{Spearman ($S$)} rank correlation against human ratings across \dataset{}, EvalCrafter, VideoFeedback, and VideoFeedback2 (for available dimensions). For feedback quality, we conduct pairwise report comparisons on \dataset{} and VBench~2.0 using Gemini~3.1~Pro~\citep{googledeepmind2026gemini31pro}. Given the video and two anonymized reports, the judge assigns \textbf{win/tie/loss/$\Delta$=win-loss} outcomes based on \textbf{(1) Accuracy} (factual correctness to the video), \textbf{(2) Groundedness} (attribution to specific, observable moments), \textbf{(3) Completeness} (coverage of prompt requirements), and \textbf{(4) Helpfulness} (clarity in explaining failure modes without rewarding verbosity), with report order swapped to mitigate positional bias. Judge prompts and human-audit results are provided in \Cref{appdx:mllm-prompts,subsec:mllm-judge-sanity} respectively.

\begin{table}[t!]
\centering
\resizebox{\textwidth}{!}{
\begin{tabular}{l | c@{\hspace{3pt}}c c@{\hspace{3pt}}c c@{\hspace{3pt}}c c@{\hspace{3pt}}c}
\toprule
& \multicolumn{2}{c}{Text-Video Align.} & \multicolumn{2}{c}{Consistency} & \multicolumn{2}{c}{Commonsense} & \multicolumn{2}{c}{Video-Audio Sync.} \\
\cmidrule(lr){2-3} \cmidrule(lr){4-5} \cmidrule(lr){6-7} \cmidrule(lr){8-9}
\textbf{Method} & P & S & P & S & P & S & P & S \\
\midrule
VBench \citep{Huang_2024_CVPR} & 0.112 & 0.181 & 0.149 & 0.176 & \cellcolor{gray!30} & \cellcolor{gray!30} & \cellcolor{gray!30} & \cellcolor{gray!30} \\
NeuS-V \citep{sharan2025neuro} & 0.263 & 0.375 & \cellcolor{gray!30} & \cellcolor{gray!30} & \cellcolor{gray!30} & \cellcolor{gray!30} & \cellcolor{gray!30} & \cellcolor{gray!30} \\
MLLM-Judge w/ SC \citep{wang2022self} & 0.360 & 0.390 & 0.182 & 0.174 & 0.240 & 0.234 & -0.110 & -0.133 \\

Self-Refine \citep{madaan2023self} & 0.333 & 0.378 & 0.230 & 0.261 & 0.168 & 0.154 & 0.189 & 0.196 \\
Multi-Agent Debate \citep{du2024improving} & 0.370 & 0.410 & 0.237 & 0.249 & 0.135 & 0.082 & 0.167 & 0.116 \\

VISTA's MMAC \citep{long2026vista} & 0.382 & 0.387 & 0.206 & 0.195 & 0.294 & 0.303 & 0.239 & 0.219 \\
VQQA \citep{song2026vqqa} & 0.408 & 0.423 & \cellcolor{gray!30} & \cellcolor{gray!30} & \cellcolor{gray!30} & \cellcolor{gray!30} & \cellcolor{gray!30} & \cellcolor{gray!30} \\
\midrule
\textbf{\model{} (w/o PAC)} & {0.550} & {0.533} & 0.364 & 0.293 & 0.326 & 0.344 & 0.657 & 0.734 \\
\textbf{\model{} (Ours)} & \textbf{0.579} & \textbf{0.566} & \textbf{0.441} & \textbf{0.452} & \textbf{0.558}  & \textbf{0.528} & \textbf{0.678} & \textbf{0.752} \\
\bottomrule
\end{tabular}
}
\caption{\textbf{Human correlation on \dataset{}.} Pearson (P) and Spearman (S) correlations across the faithfulness dimensions. \model{} significantly outperforms all baselines, with PAC providing further gains. The scores' distributions are plotted in \Cref{fig:baselines_scores_distributions}.}
\label{tab:correlation-ourdata}
\end{table}
\begin{table}[t!]
\centering
\resizebox{\linewidth}{!}{%
\begin{tabular}{l|cccc|cccc|cc}
\toprule
& \multicolumn{4}{c|}{\textbf{EvalCrafter}} & \multicolumn{4}{c|}{\textbf{VideoFeedback}} & \multicolumn{2}{c}{\textbf{VideoFeedback2}} \\
 & \multicolumn{4}{c|}{\citep{liu2024evalcrafter}} & \multicolumn{4}{c|}{\citep{he-etal-2024-videoscore}} & \multicolumn{2}{c}{\citep{he2025videoscore2}} \\
 & \multicolumn{2}{c}{Text-Video Align.} & \multicolumn{2}{c|}{Consistency} & \multicolumn{2}{c}{Text-Video Align.} & \multicolumn{2}{c|}{Consistency} & \multicolumn{2}{c}{Text-Video Align.} \\
\textbf{Method} & P & S & P & S & P & S & P & S & P & S \\
\midrule
VBench \citep{Huang_2024_CVPR} & -0.162 & -0.150 & 0.479 & 0.533 & -0.112 & -0.079 & 0.421 & 0.502 & -0.138 & -0.120 \\
NeuS-V \citep{sharan2025neuro} & 0.217 & 0.210 & \cellcolor{gray!30} & \cellcolor{gray!30} & 0.151 & 0.123 & \cellcolor{gray!30} & \cellcolor{gray!30} & 0.164 & 0.157 \\
MLLM-Judge w/ SC \citep{wang2022self} & 0.691 & 0.699 & 0.571 & 0.585 & 0.428 & 0.433 & 0.427 & 0.432 & 0.615 & 0.629 \\
Self-Refine \citep{madaan2023self} & 0.690 & 0.703 & 0.571 & 0.580 & 0.451 & 0.481 & 0.439 & 0.443 & 0.599 & 0.644 \\
Multi-Agent Debate \citep{du2024improving} & 0.653 & 0.676 & 0.534 & 0.564 & 0.484 & 0.506 & 0.402 & 0.410 & 0.592 & 0.636 \\
VISTA's MMAC \citep{long2026vista} & 0.700 & {0.708} & \textbf{0.594} & \textbf{0.596} & 0.456 & 0.504 & 0.400 & 0.418 & 0.657 & 0.671 \\
VQQA \citep{song2026vqqa} & 0.696 & 0.689 & \cellcolor{gray!30} & \cellcolor{gray!30} & 0.496 & 0.501 & \cellcolor{gray!30} & \cellcolor{gray!30} & 0.668 & 0.670 \\
\midrule
\textbf{\model{} (Ours; w/o PAC)} & 0.693 & 0.674 & 0.557 & 0.578 & 0.558 & 0.478 & 0.464 & 0.508 & 0.697 & 0.710 \\
\textbf{\model{} (Ours; w/ PAC)} & \textbf{0.734} & \textbf{0.728} & 0.578 & 0.570 & \textbf{0.564} & \textbf{0.564} & \textbf{0.502} & \textbf{0.528} & \textbf{0.705} & \textbf{0.720} \\
\bottomrule
\end{tabular}}
\caption{\textbf{Human correlation on public benchmarks.} Pearson (P) and Spearman (S) correlations for text--video alignment and temporal consistency. \colorbox{gray!30}{Gray} cells denote dimensions not supported by a baseline. 
}
\label{tab:correlation-publicdata}
\end{table}

\subsection{Human Alignment Results (RQ1)} \label{subsec:human-alignment-eval}

\paragraph{\model{} achieves superior human alignment across benchmarks.}
\Cref{tab:correlation-ourdata} outlines our correlation results on \dataset{}. \model{} outperforms the strongest baselines, particularly compared to VISTA on physical commonsense with $(P,S)$ of $(0.558,0.528)$ versus $(0.294,0.303)$ (+26\% Pearson) and video--audio synchronization of $(0.678,0.752)$  versus $(0.239,0.219)$ (+44\% Pearson), and on text--video alignment, $(0.579,0.566)$ versus $(0.408,0.423)$ of VQQA. Importantly, this superior performance generalizes to public annotations (\Cref{tab:correlation-publicdata}). For example, on EvalCrafter's text--video alignment, \model{} achieves a Pearson score of $0.734$ (versus $0.7$ for the best baseline) and $0.564$ on VideoFeedback. Our tables also show that PAC notably improves correlations on nearly all dimensions, confirming the importance of calibrating violations.

\vspace{3mm}
\noindent\textbf{Holistic MLLMs degrade notably on multi-event scenarios, failing to track conditional exceptions, invariants, bounded deadlines, and multimodal synchronization.}
On EvalCrafter and VideoFeedback2, we find that all baselines achieve $0.60$--$0.70$ Pearson correlation on text--video alignment. For EvalCrafter’s single-event prompts, MLLM baselines perform nearly on par with \model{}, while VISTA even surpasses it on consistency because its adversarial critique can identify issues in short videos without further decomposition. Meanwhile, on \dataset{} the baselines fall to only $0.33$--$0.41 (P)$. Our human investigation shows that holistic evaluators struggle with dense multi-event prompts requiring temporal logic, particularly those that involve conditional exceptions, repeated-instance constraints, global invariants, and response deadlines (e.g., actions required \emph{unless}, \emph{for each} instance, \emph{whenever}, or \emph{within a period}). See \Cref{appdx:case-studies} for some of our case studies.

\vspace{3mm}
\noindent\textbf{\model{} demonstrates robustness across evaluation dimensions.}
When analyzing human scores, we notice that their distributions naturally vary across dimensions. For example, on \dataset{} (\Cref{fig:baselines_scores_distributions}), human ratings for temporal consistency and audio--visual synchronization are quite high because SOTA video generation models already excel on them, whereas text--video alignment and physical commonsense remain challenging. We observe that existing baselines struggle to adapt to these different distributions, revealing several noticeable behavior biases. For example, MLLM evaluators tend to flatten scores across bins, VBench saturates when evaluating SOTA video models, and NeuS-V concentrates scores in the $0.0$--$0.2$ bins. For NeuS-V, this bias comes from applying strict LTL operators, such as multiplying confidences across frames for ALWAYS. This formulation is brittle, as localized MLLM error can collapse the final score, especially when evaluating many frames (doubling the \#frames drops Pearson to $\sim0.1$ on both \dataset{} and EvalCrafter). By using our soft LTL-inspired operators, \model{} prevents localized errors from dominating and better matches human-rating distributions across dimensions.

\begin{table*}[t]
\centering
\resizebox{\textwidth}{!}{%
\begin{tabular}{l|cccc|cccc|cccc|cccc|c}
\toprule
\multirow{2}{*}{\textbf{Method}} & \multicolumn{4}{c|}{\textbf{Accuracy}} & \multicolumn{4}{c|}{\textbf{Groundedness}} & \multicolumn{4}{c|}{\textbf{Completeness}} & \multicolumn{4}{c|}{\textbf{Helpfulness}} & \multirow{2}{*}{\textbf{Avg. $\Delta$}} \\
& Win & Tie & Loss & $\Delta$ & Win & Tie & Loss & $\Delta$ & Win & Tie & Loss & $\Delta$ & Win & Tie & Loss & $\Delta$ & \\
\midrule
MLLM-Judge w/ SC \citep{wang2022self} & 65.3 & 17.4 & 17.3 & +48.0 & 75.5 & 11.2 & 13.3 & +62.2 & 73.3 & 12.7 & 14.0 & +59.3 & 69.8 & 14.4 & 15.8 & +54.0 & +55.9 \\
Self-Refine \citep{madaan2023self} & 52.6 & 21.2 & 26.2 & +26.4 & 56.3 & 17.1 & 26.6 & +29.7 & 53.1 & 20.3 & 26.6 & +26.5 & 51.6 & 19.9 & 28.5 & +23.1 & +26.4 \\
Multi-Agent Debate \citep{du2024improving} & 56.3 & 19.8 & 23.9 & +32.4 & 62.4 & 16.8 & 20.8 & +41.6 & 62.3 & 15.7 & 22.0 & +40.3 & 59.5 & 18.3 & 22.2 & +37.3 & +37.9 \\
VISTA \citep{long2026vista} & 48.0 & 24.9 & 27.1 & +20.9 & 54.5 & 19.6 & 25.9 & +28.6 & 53.2 & 22.5 & 24.3 & +28.9 & 49.6 & 24.9 & 25.5 & +24.1 & +25.6 \\
VQQA \citep{song2026vqqa} & 58.0 & 20.0 & 22.0 & +36.0 & 60.4 & 21.2 & 18.4 & +42.0 & 65.9 & 17.3 & 16.8 & +49.1 & 55.9 & 21.7 & 22.4 & +33.5 & +40.2 \\
\midrule
Avg. $\Delta$ & \cellcolor{gray!30} & \cellcolor{gray!30} & \cellcolor{gray!30} & +32.7 & \cellcolor{gray!30} & \cellcolor{gray!30} & \cellcolor{gray!30} & +40.8 & \cellcolor{gray!30} & \cellcolor{gray!30} & \cellcolor{gray!30} & +40.8 & \cellcolor{gray!30} & \cellcolor{gray!30} & \cellcolor{gray!30} & +34.4 & +37.2 \\
\bottomrule
\end{tabular}%
}
\caption{\textbf{Averaged pairwise comparisons across dimensions and benchmarks.} Win/Tie/Loss/$\Delta=\mathrm{Win}-\mathrm{Loss}$ (\%) of our method versus each baseline. Full tabular results are provided in \Cref{appdx:details-feedback-results,tab:pairwise,tab:pairwise_commonsense,tab:pairwise_consistency,tab:pairwise_audio}.}
\label{tab:pairwise_overall}
\end{table*}


\subsection{Feedback Quality Results (RQ2)} \label{subsec:feedback-quality-eval}

\vspace{3mm}
\noindent\textbf{\model{} produces superior diagnostic reports, particularly in groundedness and completeness.} 
\Cref{tab:pairwise_overall} averages our results over all dimensions and benchmarks. We observe that the average $\Delta$ over baselines is largest on Groundedness and Completeness. For example, \model{} is more grounded than the baselines on \dataset{} when comparing physical commonsense reports with $\Delta = +55.4\%$ (\Cref{tab:pairwise_commonsense}). It is also more complete with $\Delta = +57.8\%$ when comparing text-video alignment reports (\Cref{tab:pairwise}). Through manual investigation, we observe that holistic evaluators tend to focus on scene-level events while often overlooking localized artifacts. For example, when a prompt requires a guard to tap a badge to enter a corridor, holistic evaluators often recognize that the guard enters the corridor, yet may overlook the ordering error if the door opens slightly before the badge is tapped. Because \model{} evaluates these implicit requirements as isolated propositions, it can reliably capture the error in time. In addition, \model{} also achieves substantial gains in Accuracy and Helpfulness with average $\Delta$ more than $30\%$, further demonstrating the quality of its diagnostic reports. We provide more examples in our supplementary materials.

\vspace{3mm}
\noindent\textbf{\model{}'s reports excel further as scenarios encode more complex temporal constraints.}
Similar to our previous observations on correlations, our pairwise margins ($\Delta$) increase substantially from VBench~2.0 to \dataset{}, jumping from $+33.2\%$ to $+47.9\%$ on alignment and from $+10.5\%$ to $+50.1\%$ on commonsense. Comparing two datasets, we find that VBench~2.0 prompts primarily require event ordering, which MLLMs and agentic frameworks can often verify well through scene-level reasoning. \dataset{}, however, introduces more complex temporal constraints (e.g., deadlines, exceptions, invariants), which demand more precise checks. This further consolidates our conclusion that a specification-driven approach is essential for evaluating complex temporal and logical dependencies.

\section{Analysis}\label{sec:main-analysis}

\subsection{Ablation Studies (RQ3)} \label{subsec:ablation-studies}

\begin{table*}[t!]
\centering
\resizebox{\linewidth}{!}{
\begin{tabular}{l cc cc cc cc}
\toprule
& \multicolumn{4}{c}{\textbf{\dataset{}}} & \multicolumn{2}{c}{\textbf{VideoFeedback} \citep{he-etal-2024-videoscore}} \\
\cmidrule(lr){2-5} \cmidrule(lr){6-7}
& Text-Video Align. & Consist. & Commonsense & Video-Audio Sync. & Text-Video Align. & Consist.\\
\midrule
\model{} w/o PAC (Ours) & \textbf{0.550 / 0.533} & \textbf{0.364 / 0.293} & \textbf{0.326 / 0.344} & \textbf{0.657 / 0.734} & \textbf{0.558} / 0.478 & \textbf{0.464 / 0.508} \\
\midrule
\model{} (Neural-only) & 0.441 / 0.437 & 0.305 / 0.286 & 0.312 / 0.327 & 0.588 / 0.640 & 0.503 / \textbf{0.525} & 0.450 / 0.484 \\
\model{} (Graph-only) & {0.085 / 0.085} & 0.149 / 0.236 & {0.088 / 0.066} & \cellcolor{gray!30}{N/A} & 0.329 / 0.324 & 0.269 / 0.247 \\
\midrule
\model{} (w/ Original LTL) & {0.091 / 0.142} & 0.113 / 0.140 & -0.089 / -0.105 & \cellcolor{gray!30} & 0.368 / 0.415 & 0.204 / 0.237 \\
\model{} (w/o Temporal Logic) & 0.183 / 0.197 & \cellcolor{gray!30} & 0.148 / 0.159 & \cellcolor{gray!30} & 0.535 / 0.530 & \cellcolor{gray!30} \\
\midrule
\model{} (Invariant only) & \cellcolor{gray!30} & \cellcolor{gray!30} & 0.188 / 0.166 & \cellcolor{gray!30} & \cellcolor{gray!30} & \cellcolor{gray!30} \\
\model{} (Inv.+Causality) & \cellcolor{gray!30} & \cellcolor{gray!30} & 0.256 / 0.230 & \cellcolor{gray!30} & \cellcolor{gray!30} & \cellcolor{gray!30} \\
\midrule
\model{} (2x frames) & 0.520 / 0.514 & 0.289 / 0.296 & 0.344 / 0.313 & \cellcolor{gray!30} & 0.557 / 0.555 & 0.520 / 0.579 \\
\model{} (4x frames) & 0.277 / 0.310 & 0.232 / 0.259 & 0.147 / 0.133 & \cellcolor{gray!30} & 0.580 / 0.586 & 0.449 / 0.514 \\
\bottomrule
\end{tabular}
}
\caption{\textbf{Ablation studies} on the core components of \textbf{\model{} w/o PAC}: tool routing, temporal verification, implicit propositions, and frame rates. We report ($P$) / ($S$) correlations. \colorbox{gray!30}{{N/A}} denotes inapplicable settings, while \colorbox{gray!30}{{-}} denotes dimensions unaffected by the ablations.}
\label{tab:ablation_tools}
\vspace{-3mm}
\end{table*}

We ablate four components of \model{} without PAC. We deliberately isolate the uncalibrated execution engine because applying PAC could obscure the raw effects of the ablations. Specifically, we study: \textbf{(i) tool routing}, comparing our full hybrid engine with \emph{Neural-only} and \emph{Graph-only}; \textbf{(ii) confidence-trace temporal operators}, compared with original LTL and no temporal operator (use ALWAYS); \textbf{(iii) implicit propositions}, incrementally adding invariant, causality, and liveness schemes; and \textbf{(iv) frame sampling}, varied by 2x and 4x. We report P/S scores on VideoFeedback and \dataset{}; gray cells indicate inapplicable settings.

\vspace{3mm}
\noindent\textbf{Hybrid routing and confidence-trace temporal reasoning are crucial.}
On \dataset{} alignment, we see that the full model achieves $0.550/0.533$, outperforming neural-only ($0.441/0.437$) and graph-only ($0.085/0.085$). Graph-only performs poorly on consistency and commonsense, where we observe frequent false negatives. This is because symbolic queries are bound to the scene graph; any visual details missed during construction evaluate to false. Meanwhile, neural-only remains competitive on simple prompts in VideoFeedback but degrades on the complex multi-event ones in \dataset{}, highlighting the importance of hybrid verification. We also see that removing temporal operators notably reduces \dataset{} scores, while the original LTL lowers them further on all metrics. These results highlight the importance of our soft operator design.

\vspace{3mm}
\noindent\textbf{Implicit propositions improve commonsense evaluation, while denser sampling introduces false violations.}
Adding invariant, causality, and liveness propositions progressively improves commonsense correlation from $0.188/0.166$ to $0.256/0.230$ approaching the full model's $0.326/0.344$. This improvement demonstrates both the importance of these proposition families and the effectiveness of deriving situation-specific constraints for detecting human-perceptible physical failures. In addition, we notice that denser sampling is not always helpful. On VideoFeedback, $4\times$ sampling reduces \dataset{} alignment to $0.277/0.310$ and commonsense to $0.147/0.133$. Our manual inspection reveals that oversampling introduces noisy intermediate observations into the scene graph. For example, rapid motion can produce blurred frames in which detectors temporarily lose objects, causing false detected violations.

\subsection{Proposition and Routing Analysis (RQ4)}

We conduct a careful human evaluation of proposition quality and tool-routing accuracy. Three highly qualified annotators evaluate propositions from 150 videos sampled across \dataset{}, VBench 2.0, and VideoFeedback2, rating \textbf{(i) Faithfulness} to the ground-truth prompt or video constraints as a binary metric and \textbf{(ii) Completeness} on a 1--5 scale \emph{for two sets of explicit and implicit propositions}. Another set of three annotators evaluates whether the assigned tool is optimal for 120 routed propositions sampled from \dataset{}.

\vspace{3mm}
\noindent\textbf{Proposition compilation and tool routing are highly reliable.}
Our results are provided in \Cref{subsec:human-verification-propositions-tools}. As shown in \Cref{tab:prop-quality,tab:tool-routing}, explicit propositions achieve $96.0\%$--$97.1\%$ faithfulness and $4.45$--$4.62$ completeness, while implicit propositions achieve $82.6\%$--$88.9\%$ and $4.09$--$4.56$, respectively. The lower implicit scores reflect the open-ended nature of deriving unstated physical constraints, for which annotators may disagree on the necessary rules. Routing achieves $100\%$ correctness for the Shot, Depth, and MLLM tools and $80\%$--$94\%$ for Motion and Audio, where general object actions, such as throwing a ball or playing an instrument, are occasionally routed to the MLLM. While our results are MLLM-specific, they show that compiling videos into formal specifications and routing them to specialized verifiers are highly tractable.

\subsection{PAC Analysis (RQ5)}

We evaluate PAC as both a micro-level classifier and macro-level calibrator using 150 PAC-excluded explicit and implicit propositions, labeled by three annotators as either imperceptible artifacts or significant failures. We report PAC's F1 and recall against their majority vote, and compare pre- versus post-calibration score distributions on \dataset{}.

\vspace{3mm}
\noindent\textbf{PAC reliably isolates human-salient errors.}
PAC achieves $85.7\%$ F1 and $90.0\%$ recall, indicating that it rarely filters out genuine generation errors. Studying false cases, we conclude that PAC’s remaining errors arise from the context-dependent nature of perceptual significance. For example, a missing object shadow might be an imperceptible artifact in a fast-moving action sequence, but that same violation becomes a glaring failure in a close-up shot. Additionally, as shown in \Cref{fig:sac_effects}, \model{}'s uncalibrated scores generally agree with human ratings but tend to over-penalize videos. PAC effectively mitigates this, preserving low scores for flawed videos while restoring high scores for successful ones and shifting the distribution closer to human judgments.

\section{Conclusions}
We introduced \model{}, a neural-symbolic specification-driven video evaluation framework that compiles prompts and videos into explicit and implicit propositions, routes them to specialized neural and graph verifiers, and evaluates their confidence traces using symbolic operators. Across public and complex multi-event benchmarks, \model{} outperforms traditional and agentic baselines in human alignment and diagnostic accuracy, groundedness, completeness, and helpfulness. These results establish proposition-level verification as a robust foundation for more human-aligned and interpretable video evaluation.

\section*{LLM Usage Statement}
The authors used ChatGPT, Gemini, and Claude to assist with coding and manuscript preparation. All substantive content, analyses, and conclusions are the authors' own, and the authors take full responsibility for the work.

\bibliographystyle{abbrvnat}
\bibliography{main}

@inproceedings{guan2025etva,
  title={Etva: Evaluation of text-to-video alignment via fine-grained question generation and answering},
  author={Guan, Kaisi and Lai, Zhengfeng and Sun, Yuchong and Zhang, Peng and Liu, Wei and Liu, Kieran and Cao, Meng and Song, Ruihua},
  booktitle={2025 IEEE/CVF International Conference on Computer Vision (ICCV)},
  pages={21299--21309},
  year={2025},
  organization={IEEE}
}

@inproceedings{zhou2026avgen,
  title={AVGen-Bench: A Task-Driven Benchmark for Multi-Granular Evaluation of Text-to-Audio-Video Generation},
  author={Zhou, Ziwei and Lai, Zeyuan and Wang, Rui and Yang, Yifan and Yang, Yuqing and Dai, Qi and Qiu, Lili and Luo, Chong},
  booktitle={Forty-third International Conference on Machine Learning},
  year={2026}
}

@inproceedings{bansal2025videophy,
  title={Videophy: Evaluating physical commonsense for video generation},
  author={Bansal, Hritik and Lin, Zongyu and Xie, Tianyi and Zong, Zeshun and Yarom, Michal and Bitton, Yonatan and Jiang, Chenfanfu and Sun, Yizhou and Chang, Kai-Wei and Grover, Aditya},
  booktitle={International Conference on Learning Representations},
  volume={2025},
  pages={102075--102121},
  year={2025}
}

@book{baier2008principles,
  title={Principles of model checking},
  author={Baier, Christel and Katoen, Joost-Pieter and Larsen, Kim Guldstrand},
  volume={735},
  year={2008},
  publisher={MIT press Cambridge}
}

@inproceedings{pnueli1977temporal,
  title={The temporal logic of programs},
  author={Pnueli, Amir},
  booktitle={18th annual symposium on foundations of computer science (sfcs 1977)},
  pages={46--57},
  year={1977},
  organization={ieee}
}

@article{he2025videoscore2,
  title={Videoscore2: Think before you score in generative video evaluation},
  author={He, Xuan and Jiang, Dongfu and Nie, Ping and Liu, Minghao and Jiang, Zhengxuan and Su, Mingyi and Ma, Wentao and Lin, Junru and Ye, Chun and Lu, Yi and others},
  journal={arXiv preprint arXiv:2509.22799},
  year={2025}
}

@inproceedings{du2024improving,
  title={Improving factuality and reasoning in language models through multiagent debate},
  author={Du, Yilun and Li, Shuang and Torralba, Antonio and Tenenbaum, Joshua B and Mordatch, Igor},
  booktitle={Proceedings of the 41st International Conference on Machine Learning},
  pages={11733--11763},
  year={2024}
}

@article{liu2023fetv,
  title={Fetv: A benchmark for fine-grained evaluation of open-domain text-to-video generation},
  author={Liu, Yuanxin and Li, Lei and Ren, Shuhuai and Gao, Rundong and Li, Shicheng and Chen, Sishuo and Sun, Xu and Hou, Lu},
  journal={Advances in Neural Information Processing Systems},
  volume={36},
  pages={62352--62387},
  year={2023}
}

@inproceedings{hessel2021clipscore,
  title={Clipscore: A reference-free evaluation metric for image captioning},
  author={Hessel, Jack and Holtzman, Ari and Forbes, Maxwell and Le Bras, Ronan and Choi, Yejin},
  booktitle={Proceedings of the 2021 conference on empirical methods in natural language processing},
  pages={7514--7528},
  year={2021}
}

@article{unterthiner2018towards,
  title={Towards accurate generative models of video: A new metric \& challenges},
  author={Unterthiner, Thomas and Van Steenkiste, Sjoerd and Kurach, Karol and Marinier, Raphael and Michalski, Marcin and Gelly, Sylvain},
  journal={arXiv preprint arXiv:1812.01717},
  year={2018}
}

@article{song2026vqqa,
  title={VQQA: An Agentic Approach for Video Evaluation and Quality Improvement},
  author={Song, Yiwen and Pfister, Tomas and Song, Yale},
  journal={arXiv preprint arXiv:2603.12310},
  year={2026}
}

@misc{seedance2,
  title        = {Seedance 2.0},
  author       = {{ByteDance Seed}},
  howpublished = {\url{https://seedance2.ai/}},
  year={2026}
}

@inproceedings{suris2023vipergpt,
  title={Vipergpt: Visual inference via python execution for reasoning},
  author={Sur{\'\i}s, D{\'\i}dac and Menon, Sachit and Vondrick, Carl},
  booktitle={2023 IEEE/CVF International Conference on Computer Vision (ICCV)},
  pages={11854--11864},
  year={2023},
  organization={IEEE}
}

@inproceedings{gupta2023visual,
  title={Visual programming: Compositional visual reasoning without training},
  author={Gupta, Tanmay and Kembhavi, Aniruddha},
  booktitle={2023 IEEE/CVF Conference on Computer Vision and Pattern Recognition (CVPR)},
  pages={14953--14962},
  year={2023},
  organization={IEEE}
}

@article{yi2020clevrer,
  title={Clevrer: Collision events for video representation and reasoning},
  author={Yi, Kexin and Gan, Chuang and Li, Yunzhu and Kohli, Pushmeet and Wu, Jiajun and Torralba, Antonio and Tenenbaum, Joshua B},
  journal={arXiv preprint arXiv:1910.01442},
  year={2019}
}

@article{mao2019neuro,
  title={The neuro-symbolic concept learner: Interpreting scenes, words, and sentences from natural supervision},
  author={Mao, Jiayuan and Gan, Chuang and Kohli, Pushmeet and Tenenbaum, Joshua B and Wu, Jiajun},
  journal={arXiv preprint arXiv:1904.12584},
  year={2019}
}

@article{yi2018neural,
  title={Neural-symbolic vqa: Disentangling reasoning from vision and language understanding},
  author={Yi, Kexin and Wu, Jiajun and Gan, Chuang and Torralba, Antonio and Kohli, Pushmeet and Tenenbaum, Josh},
  journal={Advances in neural information processing systems},
  volume={31},
  year={2018}
}

@article{liu2026improving,
  title={Improving video generation with human feedback},
  author={Liu, Jie and Liu, Gongye and Liang, Jiajun and Yuan, Ziyang and Liu, Xiaokun and Zheng, Mingwu and Wu, Xiele and Wang, Qiulin and Xia, Menghan and Wang, Xintao and others},
  journal={Advances in Neural Information Processing Systems},
  volume={38},
  pages={82155--82192},
  year={2026}
}

@article{chen2024mllm,
  title={Mllm-as-a-judge: Assessing multimodal llm-as-a-judge with vision-language benchmark},
  author={Chen, Dongping and Chen, Ruoxi and Zhang, Shilin and Liu, Yinuo and Wang, Yaochen and Zhou, Huichi and Zhang, Qihui and Wan, Yao and Zhou, Pan and Sun, Lichao},
  journal={arXiv preprint arXiv:2402.04788},
  year={2024}
}

@inproceedings{sun2024t2v,
  title={T2V-CompBench: A Comprehensive Benchmark for Compositional Text-to-video Generation},
  author={Sun, Kaiyue and Huang, Kaiyi and Liu, Xian and Wu, Yue and Xu, Zihan and Li, Zhenguo and Liu, Xihui},
  booktitle={Proceedings of the IEEE/CVF conference on computer vision and pattern recognition},
  year={2025}
}

@article{team2026qwen38,
  title={Qwen3.8-omni: Towards native omni-modal agents},
  author={Team, Qwen},
  journal={arXiv preprint arXiv:2609.25611},
  year={2026}
}

@misc{googledeepmind2026gemini31pro,
  title        = {Gemini 3.1 Pro: A smarter model for your most complex tasks},
  author       = {{Google}},
  year         = {2026},
  howpublished = {\url{https://blog.google/innovation-and-ai/models-and-research/gemini-models/gemini-3-1-pro/}},
}

@misc{googledeepmind2025gemini3flash,
  author       = {{Google}},
  title        = {{Gemini 3 Flash: frontier intelligence built for speed}},
  howpublished = {\url{"https://blog.google/products-and-platforms/products/gemini/gemini-3-flash/"}},
  year         = {2025},
}

@article{huang2025vbench,
  title={Vbench++: Comprehensive and versatile benchmark suite for video generative models},
  author={Huang, Ziqi and Zhang, Fan and Xu, Xiaojie and He, Yinan and Yu, Jiashuo and Dong, Ziyue and Ma, Qianli and Chanpaisit, Nattapol and Si, Chenyang and Jiang, Yuming and others},
  journal={IEEE Transactions on Pattern Analysis and Machine Intelligence},
  year={2025},
  publisher={IEEE}
}

@inproceedings{bansal2026videophy,
  title={Videophy-2: A challenging action-centric physical commonsense evaluation in video generation},
  author={Bansal, Hritik and Peng, Clark and Bitton, Yonatan and Goldenberg, Roman and Grover, Aditya and Chang, Kai-Wei},
  booktitle={International Conference on Learning Representations},
  volume={2026},
  pages={118456--118470},
  year={2026}
}

@inproceedings{cho2024davidsonian,
  title={Davidsonian scene graph: Improving reliability in fine-grained evaluation for text-to-image generation},
  author={Cho, Jaemin and Hu, Yushi and Baldridge, Jason and Garg, Roopal and Anderson, Peter and Krishna, Ranjay and Bansal, Mohit and Pont-Tuset, Jordi and Wang, Su},
  booktitle={International conference on learning representations},
  volume={2024},
  pages={15625--15645},
  year={2024}
}

@inproceedings{hu2023tifa,
  title={Tifa: Accurate and interpretable text-to-image faithfulness evaluation with question answering},
  author={Hu, Yushi and Liu, Benlin and Kasai, Jungo and Wang, Yizhong and Ostendorf, Mari and Krishna, Ranjay and Smith, Noah A},
  booktitle={2023 ieee/cvf international conference on computer vision (iccv)},
  pages={20349--20360},
  year={2023},
  organization={IEEE}
}

@inproceedings{long2026vista,
  title={Vista: A test-time self-improving video generation agent},
  author={Long, Do Xuan and Wan, Xingchen and Nakhost, Hootan and Lee, Chen-Yu and Pfister, Tomas and Arik, Sercan {\"O}},
  booktitle={Proceedings of the IEEE/CVF conference on computer vision and pattern recognition},
  pages={6021--6032},
  year={2026}
}

@misc{qwen38,
    title = {Qwen3.8-Max: A New Bar for Coding and Cowork},
    url = {https://qwen.ai/blog?id=qwen3.8},
    author = {{Qwen Team}},
    month = {August},
    year = {2026}
}

@article{lai2026minimax,
  title={Minimax sparse attention},
  author={Lai, Xunhao and Xu, Weiqi and Yang, Yufeng and Chen, Qiaorui and Xu, Yang and Zeng, Lunbin and Li, Xiaolong and Sun, Haohai and Zhu, Haichao and Zhang, Vito and others},
  journal={arXiv preprint arXiv:2606.13392},
  year={2026}
}

@article{long20262,
  title={A$^2$RD: Agentic Autoregressive Diffusion for Long Video Consistency},
  author={Long, Do Xuan and Song, Yale and Kan, Min-Yen and Pfister, Tomas and Le, Long T},
  journal={arXiv preprint arXiv:2605.06924},
  year={2026}
}

@article{guo2025t2vphysbench,
  title={T2vphysbench: A first-principles benchmark for physical consistency in text-to-video generation},
  author={Guo, Xuyang and Huo, Jiayan and Shi, Zhenmei and Song, Zhao and Zhang, Jiahao and Zhao, Jiale},
  journal={arXiv preprint arXiv:2505.00337},
  year={2025}
}

@article{zheng2025locot2v,
  title={LoCoT2V-Bench: Benchmarking Long-Form and Complex Text-to-Video Generation},
  author={Zheng, Xiangqing and Wu, Chengyue and Chen, Kehai and Zhang, Min},
  journal={arXiv preprint arXiv:2510.26412},
  year={2025}
}

@inproceedings{hua2026vabench,
  title={Vabench: A comprehensive benchmark for audio-video generation},
  author={Hua, Daili and Wang, Xizhi and Zeng, Bohan and Huang, Xinyi and Liang, Hao and Niu, Junbo and Chen, Xinlong and Xu, Quanqing and Zhang, Wentao},
  booktitle={Proceedings of the IEEE/CVF Conference on Computer Vision and Pattern Recognition},
  pages={23345--23355},
  year={2026}
}

@article{tilak2026brite,
  title={BRITE: A Benchmark for Reliable and Interpretable T2V Evaluation on Implausible Scenarios},
  author={Tilak, Advait and Choi, Jiwon and Mouli, Nazifa and Le, Wei},
  journal={arXiv preprint arXiv:2605.00873},
  year={2026}
}

@inproceedings{dwyer1998property,
  title={Property specification patterns for finite-state verification},
  author={Dwyer, Matthew B and Avrunin, George S and Corbett, James C},
  booktitle={Proceedings of the second workshop on Formal methods in software practice},
  pages={7--15},
  year={1998}
}

@inproceedings{han2024folio,
  title={Folio: Natural language reasoning with first-order logic},
  author={Han, Simeng and Schoelkopf, Hailey and Zhao, Yilun and Qi, Zhenting and Riddell, Martin and Zhou, Wenfei and Coady, James and Peng, David and Qiao, Yujie and Benson, Luke and others},
  booktitle={Proceedings of the 2024 Conference on Empirical Methods in Natural Language Processing},
  pages={22017--22031},
  year={2024}
}

@article{park2024sortformer,
  title={Sortformer: A novel approach for permutation-resolved speaker supervision in speech-to-text systems},
  author={Park, Taejin and Medennikov, Ivan and Dhawan, Kunal and Wang, Weiqing and Huang, He and Koluguri, Nithin Rao and Puvvada, Krishna C and Balam, Jagadeesh and Ginsburg, Boris},
  journal={arXiv preprint arXiv:2409.06656},
  year={2024}
}

@misc{aslakey_shot_scale,
  author       = {{Aslakey}},
  title        = {Shot Scale},
  howpublished = {\url{https://huggingface.co/aslakey/shot_scale}},
  publisher    = {Hugging Face},
  urldate      = {2026-09-23}
}

@article{yang2024depth,
  title={Depth anything v2},
  author={Yang, Lihe and Kang, Bingyi and Huang, Zilong and Zhao, Zhen and Xu, Xiaogang and Feng, Jiashi and Zhao, Hengshuang},
  journal={Advances in neural information processing systems},
  volume={37},
  pages={21875--21911},
  year={2024}
}

@inproceedings{carion2026sam,
  title={Sam 3: Segment anything with concepts},
  author={Carion, Nicolas and Gustafson, Laura and Hu, Yuan-Ting and Debnath, Shoubhik and Hu, Ronghang and Suris Coll-Vinent, Didac and Ryali, Chaitanya and Alwala, Kalyan Vasudev and Khedr, Haitham and Huang, Andrew and others},
  booktitle={International conference on learning representations},
  volume={2026},
  pages={138846--138923},
  year={2026}
}

@inproceedings{goncalves2024perceptual,
  title={Perceptual evaluation of audio-visual synchrony grounded in viewers’ opinion scores},
  author={Goncalves, Lucas and Mathur, Prashant and Lavania, Chandrashekhar and Cekic, Metehan and Federico, Marcello and Han, Kyu J},
  booktitle={European Conference on Computer Vision},
  pages={288--305},
  year={2024},
  organization={Springer}
}

@article{team2026qwen3,
  title={Qwen3.5-omni technical report},
  author={Team, Qwen},
  journal={arXiv preprint arXiv:2604.15804},
  year={2026}
}

@article{krippendorff1970estimating,
  title={Estimating the reliability, systematic error and random error of interval data},
  author={Krippendorff, Klaus},
  journal={Educational and psychological measurement},
  volume={30},
  number={1},
  pages={61--70},
  year={1970},
  publisher={Sage Publications Sage CA: Thousand Oaks, CA}
}

@inproceedings{
zheng2023judging,
title={Judging {LLM}-as-a-Judge with {MT}-Bench and Chatbot Arena},
author={Lianmin Zheng and Wei-Lin Chiang and Ying Sheng and Siyuan Zhuang and Zhanghao Wu and Yonghao Zhuang and Zi Lin and Zhuohan Li and Dacheng Li and Eric Xing and Hao Zhang and Joseph E. Gonzalez and Ion Stoica},
booktitle={Thirty-seventh Conference on Neural Information Processing Systems Datasets and Benchmarks Track},
year={2023},
url={https://openreview.net/forum?id=uccHPGDlao}
}

@article{lu2024surveying,
  title={Surveying neuro-symbolic approaches for reliable artificial intelligence of things},
  author={Lu, Zhen and Afridi, Imran and Kang, Hong Jin and Ruchkin, Ivan and Zheng, Xi},
  journal={Journal of Reliable Intelligent Environments},
  volume={10},
  number={3},
  pages={257--279},
  year={2024},
  publisher={Springer}
}

@article{yu2023survey,
  title={A survey on neural-symbolic learning systems},
  author={Yu, Dongran and Yang, Bo and Liu, Dayou and Wang, Hui and Pan, Shirui},
  journal={Neural Networks},
  volume={166},
  pages={105--126},
  year={2023},
  publisher={Elsevier}
}

@inproceedings{hu2016harnessing,
  title={Harnessing deep neural networks with logic rules},
  author={Hu, Zhiting and Ma, Xuezhe and Liu, Zhengzhong and Hovy, Eduard and Xing, Eric},
  booktitle={Proceedings of the 54th Annual Meeting of the Association for Computational Linguistics (Volume 1: Long Papers)},
  pages={2410--2420},
  year={2016}
}

@inproceedings{liu2024evalcrafter,
  title={Evalcrafter: Benchmarking and evaluating large video generation models},
  author={Liu, Yaofang and Cun, Xiaodong and Liu, Xuebo and Wang, Xintao and Zhang, Yong and Chen, Haoxin and Liu, Yang and Zeng, Tieyong and Chan, Raymond and Shan, Ying},
  booktitle={Proceedings of the IEEE/CVF Conference on Computer Vision and Pattern Recognition},
  pages={22139--22149},
  year={2024},
}

@misc{deepmind2025veo3,
  author       = {{Google Deepmind}},
  title        = {{Veo}},
  year         = {2025},
  howpublished = {\url{https://deepmind.google/models/veo/}},
}

@inproceedings{he-etal-2024-videoscore,
  title={Videoscore: Building automatic metrics to simulate fine-grained human feedback for video generation},
  author={He, Xuan and Jiang, Dongfu and Zhang, Ge and Ku, Max and Soni, Achint and Siu, Sherman and Chen, Haonan and Chandra, Abhranil and Jiang, Ziyan and Arulraj, Aaran and others},
  booktitle={Proceedings of the 2024 Conference on Empirical Methods in Natural Language Processing},
  pages={2105--2123},
  year={2024}
}

@inproceedings{manna1990hierarchy,
  title={A hierarchy of temporal properties (invited paper, 1989)},
  author={Manna, Zohar and Pnueli, Amir},
  booktitle={Proceedings of the ninth annual ACM symposium on Principles of distributed computing},
  pages={377--410},
  year={1990}
}

@article{salimans2016improved,
  title={Improved techniques for training gans},
  author={Salimans, Tim and Goodfellow, Ian and Zaremba, Wojciech and Cheung, Vicki and Radford, Alec and Chen, Xi},
  journal={Advances in neural information processing systems},
  volume={29},
  year={2016},
url="https://proceedings.neurips.cc/paper/2016/file/8a3363abe792db2d8761d6403605aeb7-Paper.pdf"
}

@article{heusel2017gans,
  title={Gans trained by a two time-scale update rule converge to a local nash equilibrium},
  author={Heusel, Martin and Ramsauer, Hubert and Unterthiner, Thomas and Nessler, Bernhard and Hochreiter, Sepp},
  journal={Advances in neural information processing systems},
  volume={30},
  year={2017},
url="https://proceedings.neurips.cc/paper_files/paper/2017/file/8a1d694707eb0fefe65871369074926d-Paper.pdf"
}

@article{madaan2023self,
  title={Self-refine: Iterative refinement with self-feedback},
  author={Madaan, Aman and Tandon, Niket and Gupta, Prakhar and Hallinan, Skyler and Gao, Luyu and Wiegreffe, Sarah and Alon, Uri and Dziri, Nouha and Prabhumoye, Shrimai and Yang, Yiming and others},
  journal={Advances in Neural Information Processing Systems},
  volume={36},
  pages={46534--46594},
  year={2023}
  }

@inproceedings{
wang2022self,
title={Self-Consistency Improves Chain of Thought Reasoning in Language Models},
author={Xuezhi Wang and Jason Wei and Dale Schuurmans and Quoc V Le and Ed H. Chi and Sharan Narang and Aakanksha Chowdhery and Denny Zhou},
booktitle={The Eleventh International Conference on Learning Representations },
year={2023},
}

@inproceedings{sharan2025neuro,
  title={Neuro-symbolic evaluation of text-to-video models using formal verification},
  author={Sharan, SP and Choi, Minkyu and Shah, Sahil and Goel, Harsh and Omama, Mohammad and Chinchali, Sandeep},
  booktitle={Proceedings of the IEEE/CVF Conference on Computer Vision and Pattern Recognition},
  pages={8395--8405},
  year={2025}
}

@article{zheng2025vbench,
  title={Vbench-2.0: Advancing video generation benchmark suite for intrinsic faithfulness},
  author={Zheng, Dian and Huang, Ziqi and Liu, Hongbo and Zou, Kai and He, Yinan and Zhang, Fan and Gu, Lulu and Zhang, Yuanhan and He, Jingwen and Zheng, Wei-Shi and others},
  journal={arXiv preprint arXiv:2503.21755},
  year={2025}
}

@inproceedings{Huang_2024_CVPR,
  title={Vbench: Comprehensive benchmark suite for video generative models},
  author={Huang, Ziqi and He, Yinan and Yu, Jiashuo and Zhang, Fan and Si, Chenyang and Jiang, Yuming and Zhang, Yuanhan and Wu, Tianxing and Jin, Qingyang and Chanpaisit, Nattapol and others},
  booktitle={Proceedings of the IEEE/CVF Conference on Computer Vision and Pattern Recognition},
  pages={21807--21818},
  year={2024}
}

\clearpage
\onecolumn
\appendix

\begin{minipage}{\textwidth}
\begin{figure}[H]
    \centering
    \includegraphics[width=\linewidth]{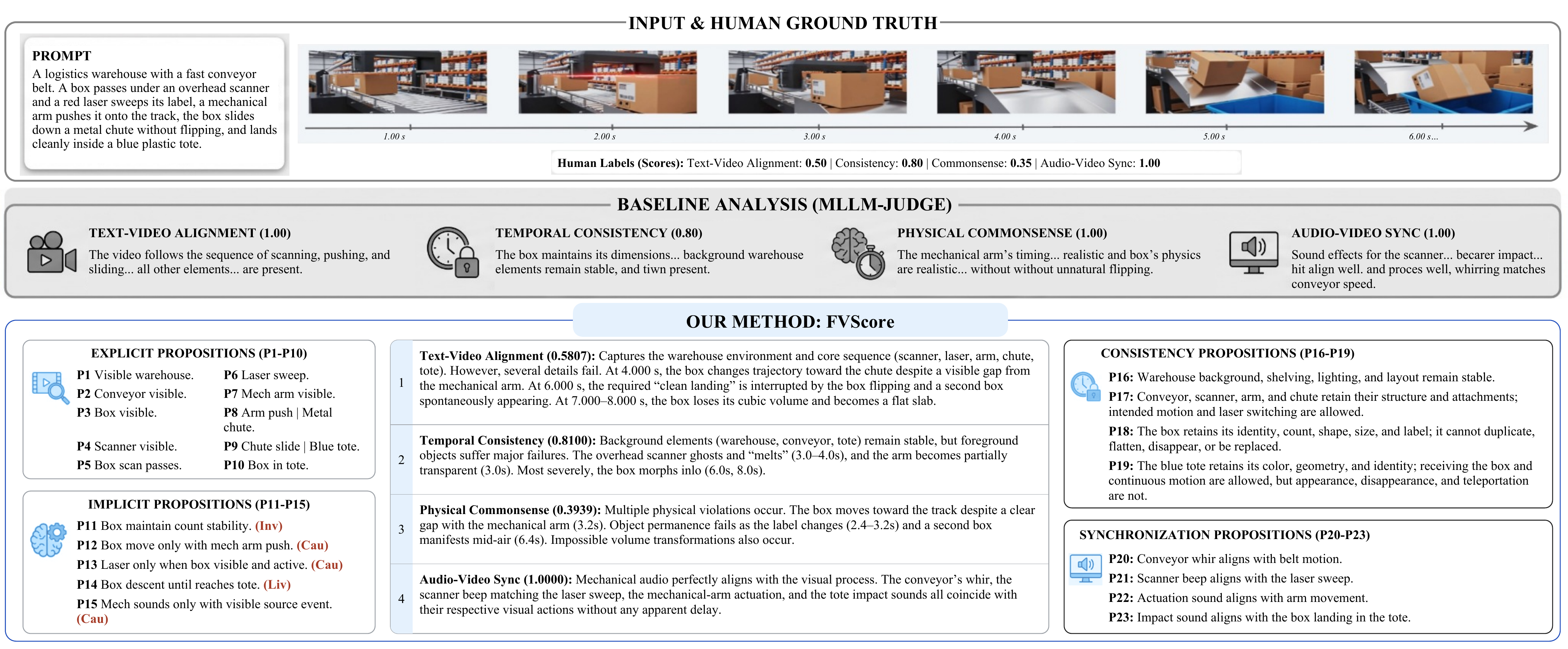}


    \includegraphics[width=\linewidth]{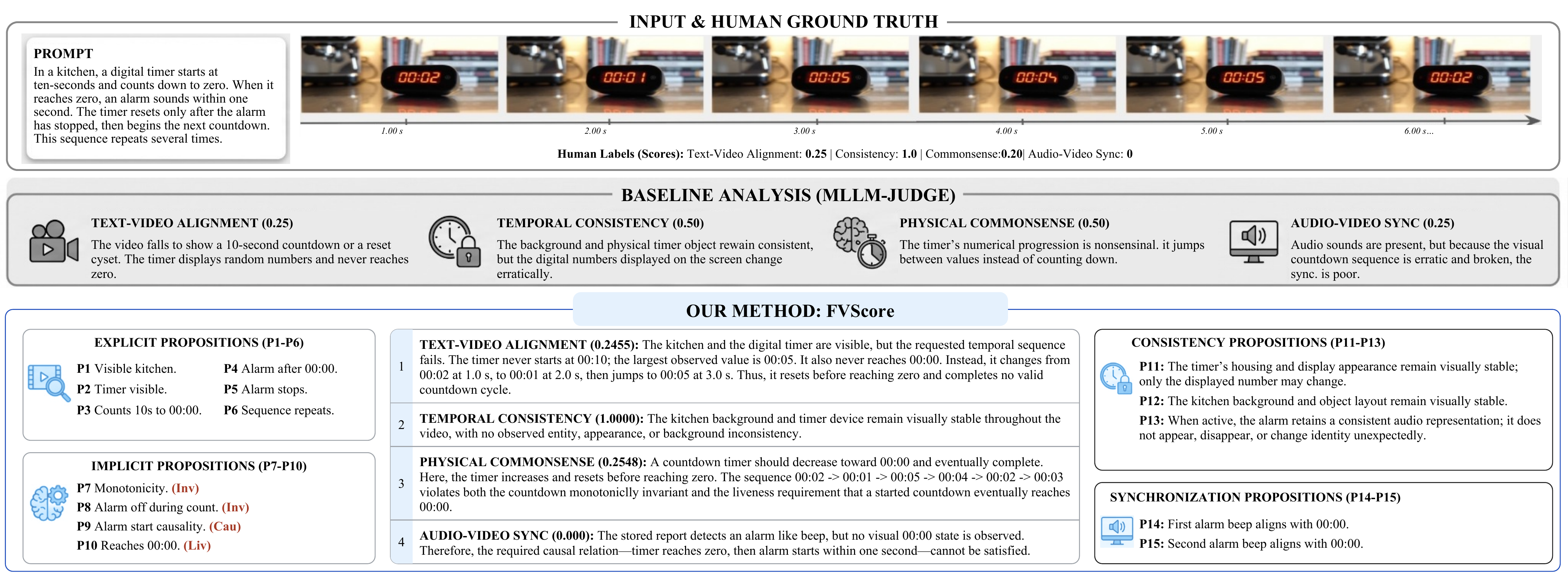}

    \caption{\textbf{Qualitative studies} between a holistic baseline and \model{}. \textbf{Top:} In the logistics scenario, the baseline hallucinates perfect physics, while \model{} catches localized artifacts (e.g., box morphing) via implicit propositions. \textbf{Bottom:} In the timer scenario, \model{} tracks monotonic countdowns and cross-modal triggers to accurately penalize logical failures where baselines fail.}
    \label{fig:qualitative_analysis}
\end{figure}
\end{minipage}

\section{Case Studies}
\label{appdx:case-studies}


We provide two qualitative examples comparing \model{} to the plain MLLM baseline illustrating how it works in \Cref{fig:qualitative_analysis}. As seen in the warehouse scenario (\Cref{fig:qualitative_analysis}, Top), the baseline MLLM assigns a perfect $1.00$ for physical commonsense, overlooking severe micro-level errors like the mechanical arm missing the box and the object flipping and unnaturally changing size by the end. By decomposing the scene into explicit and implicit propositions, \model{} successfully catches these localized physical violations—flagging exact errors in object permanence and volume conservation—to yield a score ($0.39$) that closely matches human judgment ($0.35$). Furthermore, explicit proposition tracking prevents failures on strict temporal and cross-modal rules. In the timer scenario (\Cref{fig:qualitative_analysis}, Bottom), the prompt requires a monotonic countdown followed by an alarm. While the baseline gives vague, overly forgiving scores, \model{} evaluates precise parametric constraints. It logs the exact sequence breakdown ($00:02 \rightarrow 00:01 \rightarrow 00:05$) and drops the audio--video sync score to $0.00$ upon detecting an ungrounded alarm sound, demonstrating vastly superior diagnostic accuracy. 
{Additional evaluation reports and an end-to-end example of the full \model{} pipeline are provided in our project page.}

\begin{table*}[th!]
\centering
\resizebox{\textwidth}{!}{
\begin{tabular}{l c cccc cc}
\toprule
& & \multicolumn{4}{c}{\textbf{\dataset{}}}
& \multicolumn{2}{c}{\textbf{VideoFeedback} \citep{he-etal-2024-videoscore}} \\
\cmidrule(lr){3-6} \cmidrule(lr){7-8}
\textbf{Backbone MLLM} & \textbf{PAC}
& Text-Video Align. & Consist. & Commonsense & Video-Audio Sync.
& Text-Video Align. & Consist. \\
\midrule

\textit{Gemini 3 Flash} & No
& {0.550} / 0.533
& {0.364} / 0.293
& {0.326} / 0.344
& {0.657} / 0.734
& {0.558} / 0.478
& {0.464} / 0.508 \\
\textit{Gemini 3 Flash} & Yes
& \textbf{0.579 / 0.566}
& \textbf{0.441 / 0.452}
& 0.558 / \textbf{0.528}
& \textbf{0.678 / 0.752}
& 0.564 / \textbf{0.564}
& {0.502 / 0.528} \\

\midrule
Qwen3.8-Omni & No & 0.434 / 0.492 & 0.258 / 0.307 & 0.282 / 0.297 & 0.378 / 0.415 & 0.503 / 0.523 & 0.446 / 0.498 \\

Qwen3.8-Omni & Yes & 0.519 / 0.508 & 0.341 / 0.364 & 0.415 / 0.446 & 0.462 / 0.480 & 0.532 / 0.509 & 0.542 / 0.610 \\

\midrule
Qwen3.5-9B & No
& 0.387 / 0.430 & 0.242 / 0.294 & 0.191 / 0.169 & \cellcolor{gray!30} & 0.509 / 0.506 & 0.473 / 0.500 \\

Qwen3.5-9B & Yes
& 0.458 / 0.486 & 0.303 / 0.171 & 0.270 / 0.259 & \cellcolor{gray!30} & 0.498 / 0.463 & \textbf{0.570 / 0.641} \\

\midrule
Qwen3.5-27B & No
& 0.271 / 0.228 & 0.210 / 0.156 & 0.390 / 0.401 & \cellcolor{gray!30} & 0.526 / 0.443 & 0.421 / 0.375 \\
Qwen3.5-27B & Yes
& 0.273 / 0.224 & 0.281 / 0.367 & 0.354 / 0.294 & \cellcolor{gray!30} & \textbf{0.587} / 0.518 & 0.459 / 0.416 \\

\midrule
MiniMax-M3 & No
& 0.336 / 0.375 & 0.228 / 0.303 & 0.380 / 0.364 & \cellcolor{gray!30} & 0.539 / 0.554 & 0.312 / 0.401 \\
MiniMax-M3 & Yes
& 0.385 / 0.432 & 0.241 / 0.232 & \textbf{0.560} / 0.499 & \cellcolor{gray!30} & 0.474 / 0.454 & 0.399 / 0.386 \\

\bottomrule
\end{tabular}
}
\caption{{\model{} results using different backbone MLLMs}. Each cell reports Pearson/Spearman correlation with human judgments. Gemini 3 Flash is our default.}
\label{tab:diff-backbone-mllms}
\end{table*}

\section{Limitations}
While \model{} makes substantial progress toward human-aligned video evaluation, several limitations remain that motivate future research. First, our soft temporal operators trade strict logical satisfaction for robustness to perceptual evidence. Although this design improves human correlation in our experiments notably, averaging confidence traces can occasionally underweight localized, catastrophic violations that require strict symbolic verification. Future work could explore adaptive temporal operators and severity-aware aggregation to enforce hard constraints while preserving robustness to frame-level understanding noise. Second, \model{}'s verification is constrained by the accuracy of its underlying MLLM and perceptual grounding tools during state extraction and predicate verification. However, because \model{} is modular and tool-agnostic, it will naturally benefit as foundation models and specialized perception tools continue to advance. Finally, while PAC effectively filters out perceptually negligible violations using lightweight text-level adjudication, modeling visual context-dependent significance to humans remains an open challenge. Future work could investigate visual-context-aware calibration to better capture nuanced human judgments across diverse scenarios.

\section{Additional Discussions}\label{appdx:additional-results}

\subsection{Experiments With Other MLLMs} \label{appdx:experiments-other-mllms}

\begin{table*}[th!]
\centering

\begin{subtable}[t]{0.6\textwidth}
\centering
\caption{Comparison with learned video evaluator.}
\resizebox{\linewidth}{!}{
\begin{tabular}{l c cc}
\toprule
& \multicolumn{1}{c}{\textbf{VideoFeedback}}
& \multicolumn{2}{c}{\textbf{\dataset{}}} \\
\cmidrule(lr){2-2} \cmidrule(lr){3-4}
\textbf{Method}
& \textbf{Text--Video Align.}
& \textbf{Text--Video Align.}
& \textbf{Commonsense} \\
\midrule
VideoScore2
& 0.582 / 0.579
& 0.349 / 0.365 & 0.303 / 0.289 \\
\textbf{\model{} (Ours)}
& 0.564 / 0.564
& 0.579 / 0.566 & 0.558 / 0.528 \\
\bottomrule
\end{tabular}
}
\label{tab:additional-videoscore2}
\end{subtable}
\hfill
\begin{subtable}[t]{0.36\textwidth}
\centering
\caption{Comparison with specialized video--audio evaluator.}
\resizebox{\linewidth}{!}{
\begin{tabular}{lc}
\toprule
\textbf{Method}
& \textbf{Video--Audio Sync.} \\
\midrule
PEAVS
& 0.247 / 0.272 \\
\textbf{\model{} (Ours)}
& 0.678 / 0.752 \\
\bottomrule
\end{tabular}
}
\label{tab:additional-audio-baseline}
\end{subtable}
\caption{{Additional baseline comparisons.} Each cell reports Pearson / Spearman correlation with human judgments. The video--audio comparison is conducted on \dataset{}.}
\label{tab:additional-baselines}
\end{table*}

\Cref{tab:diff-backbone-mllms} evaluates whether \model{} generalizes across different backbone MLLMs. We experimented with four other SOTA omni models across scales that are capable of understanding videos, namely \textbf{Qwen3.8-Omni} \citep{team2026qwen38}, \textbf{Qwen3.5 (9B, 27B)} \citep{team2026qwen3} and \textbf{MiniMax-M3} \citep{lai2026minimax}. 

We observe that all tested backbones produce positive to strong correlations with human judgments across the evaluated dimensions, indicating that the specification-driven framework is not specific to just Gemini models. Gemini~3~Flash nevertheless provides the strongest and most consistent overall performance, particularly on \dataset{}. Other backbones remain competitive on individual dimensions: Qwen3.5-9B surprisingly achieves the strongest VideoFeedback consistency correlation, Qwen3.5-27B performs best in Pearson correlation on VideoFeedback alignment, and MiniMax-M3 matches Gemini~3~Flash on \dataset{} physical commonsense. Nevertheless, Qwen3.8-Omni performs strongest across these newly experimented omni models.

In addition, PAC generally improves performance across backbones, increasing Pearson correlation in 12 of the 15 non-default backbone-dimension settings. However, the gains are not uniform: calibration occasionally reduces correlation, particularly on VideoFeedback alignment and for Qwen3.5-27B on physical commonsense. This variation is expected because PAC relies on the backbone's ability to distinguish perceptually significant violations from minor artifacts. Overall, the results demonstrate that \model{} transfers across both open- and closed-source MLLMs, while stronger backbone reasoning yields more reliable proposition compilation, verification, and calibration.

\subsection{Additional Baseline Comparisons}
\label{appdx:additional-baselines}

\begin{table*}[th!]
\centering
\resizebox{\textwidth}{!}{%
\begin{tabular}{@{}l cccc cc@{}}
\toprule
& \multicolumn{4}{c}{\textbf{\dataset{}}}
& \multicolumn{2}{c}{\textbf{VideoFeedback} \citep{he-etal-2024-videoscore}} \\
\cmidrule(lr){2-5} \cmidrule(lr){6-7}
\textbf{Calibration methods}
& Text-Video Align. & Consist. & Commonsense & Video-Audio Sync.
& Text-Video Align. & Consist. \\
\midrule
No calibration
& 0.550 / 0.533 & 0.364 / 0.293 & 0.326 / 0.344 & 0.657 / 0.734
& 0.558 / 0.478 & 0.464 / 0.508 \\
Random removal
& 0.512 / 0.498 & 0.341 / 0.285 & 0.368 / 0.371 & 0.589 / 0.662
& 0.519 / 0.452 & 0.381 / 0.409 \\
LLM zero-shot 
& 0.562 / 0.549 & 0.405 / 0.412 & 0.464 / 0.451 & 0.665 / 0.741
& 0.526 / 0.504 & 0.483 / 0.512 \\
\midrule
\textbf{PAC (Ours)}
& 0.579 / 0.566 & 0.441 / 0.452 & 0.558 / 0.528 & 0.678 / 0.752
& 0.564 / 0.564 & 0.502 / 0.528 \\
\bottomrule
\end{tabular}%
}
\caption{Effect of proposition removal on \model{} with Gemini~3 Flash. Each cell reports Pearson/Spearman correlation with human judgments.}
\label{tab:pac-removal-baselines}
\end{table*}

In this section, we further compare \model{} with video evaluators trained on human annotations and those specialized for individual evaluation dimensions.

We first compare \model{} with \textbf{VideoScore2}~\citep{he2025videoscore2}, a learned video evaluator, on text--video alignment and physical commonsense. On VideoFeedback, VideoScore2 slightly outperforms \model{} on alignment, reaching $0.582/0.579$ compared with $0.564/0.564$. However, on the more complex multi-event videos in \dataset{}, \model{} substantially outperforms VideoScore2 on both alignment ($0.579/0.566$ vs.\ $0.349/0.365$) and physical commonsense ($0.558/0.528$ vs.\ $0.303/0.289$). These results suggest that learned holistic scoring transfers well to simpler public videos but struggles with dense temporal and implicit physical constraints, for which proposition-level verification is particularly beneficial.

We also compare our method with \textbf{PEAVS}~\citep{goncalves2024perceptual}, a specialized reference-free metric for video--audio synchronization. \model{} achieves substantially higher correlation with human judgments ($0.678/0.752$ vs.\ $0.247/0.272$). PEAVS scores exhibit a roughly unimodal distribution concentrated between $3.0$ and $4.0$ on its five-point scale, providing limited separation among the predominantly high-quality generations in \dataset{}. Moreover, PEAVS is trained primarily to recognize generic synchrony distortions, such as temporal shifts, speed changes, muting, and segment shuffling, whereas \dataset{} requires semantic, event-specific correspondence between sounds and their visual sources. By explicitly verifying each detected sound against the concurrent visual event, \model{} better captures these fine-grained cross-modal constraints.

To isolate the contribution of \textbf{PAC}, we compares it against two alternative calibration strategies using the same Gemini~3 Flash backbone in Table~\ref{tab:pac-removal-baselines}: \textbf{Random removal}, which randomly drops propositions at the same average pruning rate as PAC (see \Cref{tab:proposition-routing}) to verify that gains do not stem merely from reducing proposition count; and \textbf{LLM zero-shot}, which prompts the same MLLM adjudicator to classify perceptual significance without held-out human-rated demonstrations. We see that random removal generally degrades correlation across dimensions by discarding genuine violations, while LLM zero-shot generally improves over uncalibrated aggregation, particularly on physical commonsense ($0.464/0.451$ vs.\ $0.326/0.344$ on \dataset{} but underperforms PAC across most settings. In summary, these experiments justify our approach of incorporating many-shot demonstrations of negligible versus human-salient violations yields the strongest alignment on both \dataset{} and VideoFeedback.

\subsection{Sanity Check on MLLM Judges}\label{subsec:mllm-judge-sanity}

\begin{table}[t!]
\centering
\resizebox{.7\linewidth}{!}{%
\begin{tabular}{lccccc}
\toprule
\textbf{Judge} & Acc. & Ground. & Compl. & Help. & \textbf{Avg.} \\
\midrule
Gemini 3.1 Pro \citep{googledeepmind2026gemini31pro}   & 74.0\% & 80.7\% & 80.7\% & 80.0\% & \textbf{78.8\%} \\
Minimax-M3 \citep{lai2026minimax}  &  66.0\% & 76.7\% & 78.0\% & 76.7\% & 74.3\% \\
Qwen 3.8 Max (0902) \citep{qwen38} & 70.0\% & 72.0\% & 75.3\% & 76.0\% & 73.3\% \\
\bottomrule
\end{tabular}}
\caption{Judge agreement with human majority vote ($\%$, 3-way Win/Tie/Loss). Human agreement is each annotator against the majority of the other two. 
}
\label{tab:judge-agreement}
\end{table}

\paragraph{Cross-Check MLLM Judge with Humans and Other MLLMs.} 
To assess the reliability of automated feedback evaluation, we cross-check three frontier MLLMs with human annotators on a random subset of 150 pairwise comparisons. For each comparison, annotators reviewed the prompt, video, and two feedback reports. They selected the better report or marked a tie separately for accuracy, groundedness, completeness, and helpfulness. For each criterion, we compute agreement as the percentage of comparisons for which the MLLM’s choice matches the human majority vote. As shown in \Cref{tab:judge-agreement}, Gemini~3.1~Pro achieves the highest agreement among the three MLLMs, averaging 78.8\% across the four criteria. We therefore use it as the primary judge for all subsequent qualitative feedback comparisons.

\paragraph{Length-Normalized Sanity Check.}

\begin{table}[t]
\centering
\resizebox{.7\textwidth}{!}{%
\begin{tabular}{lrr}
\toprule
Method & Average tokens & Average tokens (normalized) \\
\midrule
MLLM Judge & 210.38 & 431.18 \\
VISTA \citep{long2026vista} & 324.18 & 428.43 \\
Multi-Agent Debate \citep{du2024improving} & 183.21 & 381.21\\
Self-Refine \citep{madaan2023self} & 248.20 &  461.76 \\
VQQA \citep{song2026vqqa} & 123.11 &  344.90 \\
\midrule
\model{} (Ours) & 459.91 & 459.91  \\
\bottomrule
\end{tabular}}
\caption{Average length of saved evaluation reports across methods.}
\label{tab:report-length}
\end{table}

\begin{table}[t]
\centering
\begin{minipage}[t]{0.48\textwidth}
\centering
\resizebox{\linewidth}{!}{%
\begin{tabular}{l|cccc}
  \toprule
  Ann. & Faith.(Exp.) & Comp.(Exp.) & Faith.(Imp.) & Comp.(Imp.) \\
  \midrule
  1  & 97.1\%  & 4.45 & 88.9\%  & 4.09 \\
  2  & 96.8\%  & 4.53 & 84.4\%  & 4.56 \\
  3  &  96.0\%   & 4.62    & 82.6\%   & 4.33     \\
  \midrule
  Kripp. $\alpha$ &  0.64   & 0.43    & 0.44   & 0.36     \\
  \bottomrule
\end{tabular}%
}
\caption{\small Human evaluation of explicit and implicit proposition faithfulness and completeness.}
\label{tab:prop-quality}
\end{minipage}%
\hfill%
\begin{minipage}[t]{0.48\textwidth}
\centering
\resizebox{\linewidth}{!}{%
\begin{tabular}{l|ccccc}
  \toprule
  Ann. & Shot tool & Depth tool & Motion tool & Audio tool & MLLM \\
  \midrule
  1 & 100\% & 100\% & 83.3\% & 90.0\% & 100\% \\
  2 & 100\% & 100\% & 89.2\% & 94.0\% & 100\% \\
  3 & 100\% & 100\% & 80.0\% & 94.0\% & 100\% \\
  \midrule
  Kripp. $\alpha$ & 1.00 & 1.00 & 0.56 & 0.56 & 1.00 \\
  \bottomrule
\end{tabular}%
}
\caption{\small Human evaluation of tool routing accuracy across specialized verifiers.}
\label{tab:tool-routing}
\end{minipage}
\end{table}

\begin{table*}[t!]
\centering
\begin{subtable}{\textwidth}
\centering
\caption{Pairwise comparison results on \dataset{} (Ours).}
\resizebox{\textwidth}{!}{%
\begin{tabular}{l|ccc c|ccc c|ccc c|ccc c|c}
\toprule
\multirow{2}{*}{\textbf{Method}}
& \multicolumn{4}{c|}{\textbf{Accuracy}}
& \multicolumn{4}{c|}{\textbf{Groundedness}}
& \multicolumn{4}{c|}{\textbf{Completeness}}
& \multicolumn{4}{c|}{\textbf{Helpfulness}}
& \multirow{2}{*}{\textbf{Avg. $\Delta$}} \\
& Win & Tie & Loss & $\Delta$
& Win & Tie & Loss & $\Delta$
& Win & Tie & Loss & $\Delta$
& Win & Tie & Loss & $\Delta$ & \\
\midrule
MLLM-Judge
& 62.9 & 12.7 & 24.4 & +38.5
& 66.5 & 18.5 & 15.0 & +51.5
& 75.2 & 14.2 & 10.6 & +64.6
& 73.1 & 18.5 & 8.4 & +64.7
& +54.8 \\
Self-Refine
& 44.2 & 19.2 & 36.6 & +7.6
& 52.7 & 16.7 & 30.6 & +22.1
& 49.5 & 15.8 & 34.7 & +14.8
& 45.5 & 15.8 & 38.7 & +6.8
& +12.8 \\
Multi-Agent Debate
& 55.0 & 11.7 & 33.3 & +21.7
& 57.5 & 15.8 & 26.7 & +30.8
& 59.2 & 13.3 & 27.5 & +31.7
& 62.5 & 13.3 & 24.2 & +38.3
& +30.6 \\
VISTA & 46.7 & 21.5 & 31.8 & +14.9
& 50.3 & 18.6 & 31.1 & +19.2
& 53.8 & 12.4 & 33.8 & +20.0
& 49.1 & 16.7 & 34.2 & +14.9
& +17.3 \\
VQQA
& 70.0 & 13.8 & 16.2 & +53.8
& 79.7 & 13.0 & 7.3 & +72.4
& 77.7 & 13.1 & 9.2 & +68.5
& 74.2 & 10.5 & 15.3 & +58.9
& +63.4 \\
\midrule
Avg. $\Delta$
& \cellcolor{gray!30} & \cellcolor{gray!30} & \cellcolor{gray!30} & +27.3
& \cellcolor{gray!30} & \cellcolor{gray!30} & \cellcolor{gray!30} & +39.2
& \cellcolor{gray!30} & \cellcolor{gray!30} & \cellcolor{gray!30} & +39.9
& \cellcolor{gray!30} & \cellcolor{gray!30} & \cellcolor{gray!30} & +36.7
& +35.8 \\
\bottomrule
\end{tabular}%
}
\end{subtable}
\caption{{Text-Video Alignment} results computed from the \textbf{length-normalized} pairwise results. Win/Tie/Loss/$\Delta$ = Win - Loss (\%) of our method vs.\ baselines.}
\label{tab:pairwise-with-length-normalized-baselines}
\end{table*}
  
Next, we study whether our MLLM judge exhibits length bias. \Cref{tab:report-length} reports the average lengths of the final evaluation reports, measured using NLTK tokenization. We observe that \model{} produces longer reports than the baselines, although they remain relatively concise at approximately 460 tokens. To determine whether this length difference contributes to our gains, we conduct a length-normalized sanity check using our Gemini~3.1~Pro judge.

Specifically, we rerun all methods on \dataset{} with an explicit instruction to produce final reports of approximately 500 tokens, then repeat the pairwise evaluation using the same judging protocol. We focus this sanity check on Text--Video Alignment, for which the largest number of baselines can provide feedback. As shown in \Cref{tab:pairwise-with-length-normalized-baselines}, although the win--loss margins ($\Delta$) decrease against most baselines compared with panel (b) of \Cref{tab:pairwise}, \model{} retains substantial gains across all baselines. These results suggest that the benefits of \model{} are notable and consistent beyond report length.

\subsection{Human Evaluations on Claims in Final Reports} 
We conduct a human study on the accuracy of claims made by \model{} and MLLM-Judge with two independent PhD candidate annotators. For each report, we treat each sentence as one claim and assess it against the prompt and its corresponding video. A sentence is supported when the annotator sees that the video accurately shows evidence of its assertions; otherwise, it is unsupported. We randomly sample 200 claims made by each of \model{} and MLLM-Judge, equally sampled across 4 dimensions.

Across dimensions, both annotators agreed that 178 of 200 claims (89.0\%) from \model{} and 76 of 200 claims (38.0\%) from MLLM-Judge were supported by the video evidence. These results show that \model{} produced much more accurate reports than MLLM-Judge, highlighting the strength of our neuro-symbolic approach.

\subsection{Human Evaluations on Tool Routings} \label{subsec:human-verification-propositions-tools}

\Cref{tab:prop-quality,tab:tool-routing} report annotator-level evaluations of proposition faithfulness, completeness, and routing accuracy across specialized verifiers. 

For proposition quality, three annotators reviewed the prompt, video, and compiled propositions for each of 150 sampled videos. They judged each explicit proposition as faithful if it accurately represented a requirement stated in the prompt. For each implicit proposition, they judged whether it expressed a constraint that should apply to the video regardless of whether the video violated or satisfied it. Annotators then rated the completeness of the explicit and implicit proposition sets separately on a 1--5 scale, with higher scores indicating better coverage.

For routing accuracy, a separate group of three annotators reviewed propositions sampled from FVBench. For each proposition, they judged whether the assigned verifier was the most suitable choice among the available verifiers, given the evidence needed to check the proposition. Each annotator recorded a binary judgment.

We measure inter-annotator agreement using Krippendorff's $\alpha$ \citep{krippendorff1970estimating}, with nominal $\alpha$ for binary judgments and interval $\alpha$ for 1--5 ratings.

\subsection{Cost Analysis}
We quantify \model{} cost by counting the number of MLLM calls consumed per video. On \dataset{} and VideoFeedback, \model{} with our default setting consumes an average of \textbf{23} calls per video: 14 for specification compilation and representation construction, 7 for proposition grounding, and 2 for PAC adjudication and report generation. Graph-query execution, temporal aggregation, and final score computation are purely symbolic and require no additional MLLM calls. To minimize overhead during grounding, we batch propositions across keyframes instead of calling a separate MLLM query for each proposition--frame pair. As of November 2026, with Gemini 3 Flash \citep{googledeepmind2025gemini3flash} as the MLLM, \model{} incurs an average inference cost below 0.10 USD per video.

\subsection{Proposition Counts, Routing Frequencies, and PAC Removal Frequencies}

\begin{table*}[th!]
\centering
\small
\resizebox{\textwidth}{!}{%
\begin{tabular}{@{}l rrrr rrrr rrrr@{}}
\toprule
& \multicolumn{4}{c}{\textbf{FVBench}}
& \multicolumn{4}{c}{\textbf{VBench~2.0}}
& \multicolumn{4}{c}{\textbf{VideoFeedback}} \\
\cmidrule(lr){2-5} \cmidrule(lr){6-9} \cmidrule(lr){10-13}
\textbf{Dimension}
& Avg. & Graph & Neural & PAC Rem. 
& Avg. & Graph & Neural & PAC Rem.
& Avg. & Graph & Neural & PAC Rem. \\
\midrule
Text--Video Alignment
& 9.36 & 10.1\% & 89.9\% & 6.0\%
& 7.50 & 14.9\% & 85.1\% & 4.5\%
& 4.06 & 33.6\% & 66.4\% & 10.5\%\\
Temporal Consistency
& 6.39 & 0.0\% & 100\% & 16.8\%
& 8.61 & 0.0\% & 100\% & 10.1\%
& 2.96 & 0.0\% & 100\% & 23.1\% \\
Physical Commonsense
& 13.43 & 4.5\% & 95.5\% & 24.7\%
& 12.59 & 8.7\% & 91.3\% & 28.8\%
& 21.78 & 10.7\% & 89.3\% & 56.2\%\\
Video--Audio Synchronization
& 9.05 & 0.0\% & 100\% & 15.0\%
& \cellcolor{gray!30} & \cellcolor{gray!30} & \cellcolor{gray!30} & \cellcolor{gray!30}
& \cellcolor{gray!30} & \cellcolor{gray!30} & \cellcolor{gray!30} & \cellcolor{gray!30} \\
\bottomrule
\end{tabular}}
\caption{Proposition counts and routing frequencies \textbf{before PAC} on FVBench, VBench 2.0, and VideoFeedback. We also record \textbf{PAC's percentage of removed propositions} (with Gemini 3 Flasg) to further understand its behavior.}
\label{tab:proposition-routing}
\end{table*}

Table~\ref{tab:proposition-routing} reports the mean number of propositions \model{} compiles per video and the percentage assigned to each verification route \textbf{before PAC}. We observe that physical commonsense has the most propositions per video on all three benchmarks, reaching 21.78 on VideoFeedback. In addition, graph queries are most frequent for text--video alignment, accounting for 10.1\% of its propositions on FVBench, 14.9\% on VBench~2.0, and 33.6\% on VideoFeedback. For PAC calibration, we see that removal rates are quite small for explicit text--video alignment (4.5\%--10.5\%). Meanwhile, PAC filters out substantially more propositions in physical commonsense (24.7\%--56.2\%), temporal consistency (10.1\%--23.1\%), and video-audio synchonization up to 15\%, pruning redundant or unecessary micro-rules generated from open-ended video dynamics, most notably on VideoFeedback, where 56.2\% of the 21.78 compiled commonsense propositions are removed.

\subsection{Distribution of Scores Before and After PAC}

\begin{figure*}[t!]
\centering    \includegraphics[width=\linewidth]{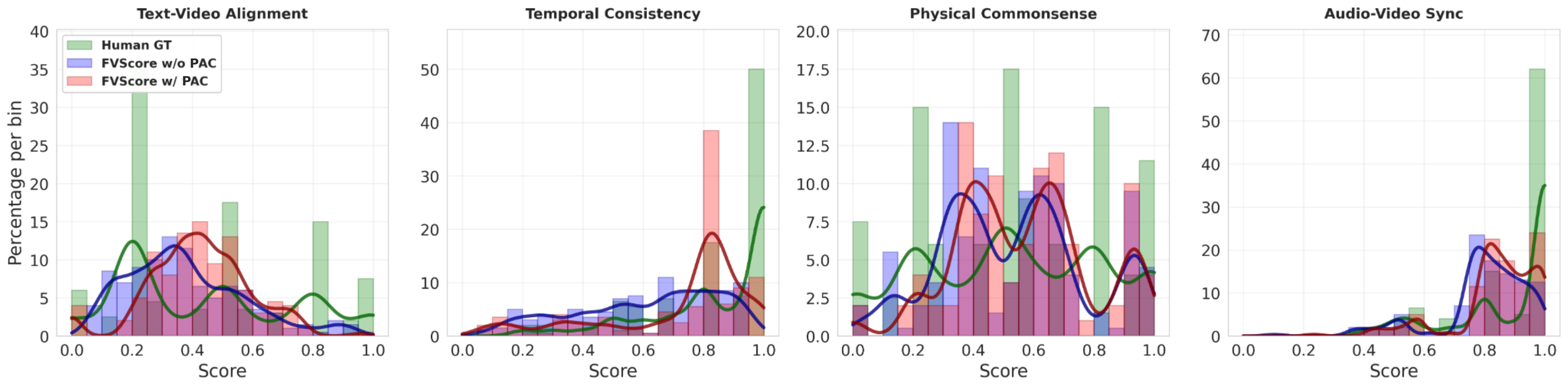}
\vspace{-5mm}
\caption{Distribution of video scores on \dataset{} before (blue) and after (red) applying PAC, compared to human ratings (green).}
\label{fig:sac_effects}
\end{figure*}

\Cref{fig:sac_effects} shows that uncalibrated verification tends to over-penalize videos, shifting scores below human ratings. PAC effectively mitigates this bias, moving the distribution closer to human judgments while preserving low scores for visibly flawed videos.

\subsection{Additional Observations About Feedback Quality} \label{appdx:additional-feedback-findings}

We observe that agentic decomposition is more effective for diagnostic critique than free-form reasoning. Across settings, we find that the plain MLLM judge has the largest gap to \model{} in six of them. While self-refinement and multi-agent debate reduce this gap, the agentic baselines remain our closest competitors in five settings: VQQA on VBench~2.0 alignment, and VISTA on consistency, commonsense, and video--audio sync. In particular, VISTA approaches parity on VBench~2.0 consistency, with a $\Delta$ of only $+3.7$ on Accuracy and $+3.6$ on Helpfulness. These results suggest agentic methods decomposing what to check work better than simply asking an MLLM to debate or critique over all dimensions and modalities. \model{} builds on this principle by using separate specialized tools for evaluations.

\subsection{Full MLLM-Judge Feedback Results} \label{appdx:details-feedback-results}

\begin{figure*}[t!]
\centering    \includegraphics[width=\linewidth]{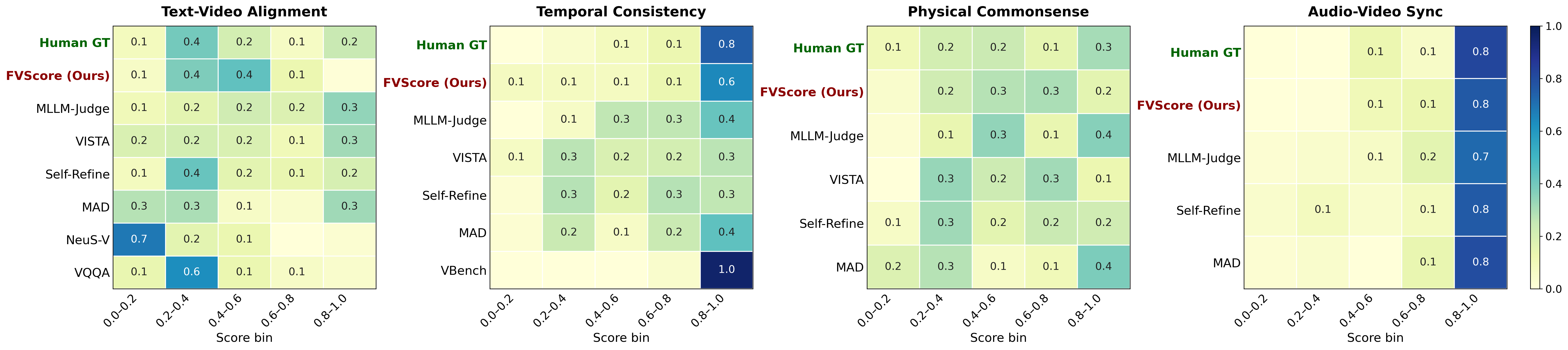}
\vspace{-5mm}
\caption{Score distributions on \dataset{}.}
\label{fig:baselines_scores_distributions}
\end{figure*}

We provide the complete pairwise evaluation of diagnostic feedback across accuracy, groundedness, completeness, and helpfulness in \Cref{tab:pairwise,tab:pairwise_consistency,tab:pairwise_commonsense,tab:pairwise_audio}. Across all dimensions, datasets, criteria, and baselines, \model{} achieves positive net win margins. On VBench~2.0, the average margins are $+33.2$ for text--video alignment, $+37.1$ for temporal consistency, and $+10.5$ for physical commonsense. The gains become particularly pronounced on \dataset{}, reaching $+47.9$ for alignment, $+36.2$ for consistency, $+50.1$ for commonsense, and $+40.1$ for video--audio synchronization. The strongest improvements generally occur in groundedness and completeness, demonstrating the value of proposition-level, evidence-grounded diagnostics.

\begin{table*}[t!]
\centering
\subfloat[Pairwise comparison results on Vbench 2.0 dataset \citep{zheng2025vbench}.]{
\resizebox{\textwidth}{!}{
\begin{tabular}{l|ccc c|ccc c|ccc c|ccc c|c}
\toprule
\multirow{2}{*}{\textbf{Method}} & \multicolumn{4}{c|}{\textbf{Accuracy}} & \multicolumn{4}{c|}{\textbf{Groundedness}} & \multicolumn{4}{c|}{\textbf{Completeness}} & \multicolumn{4}{c|}{\textbf{Helpfulness}} & \multirow{2}{*}{\textbf{Avg. $\Delta$}} \\
& Win & Tie & Loss & $\Delta$ & Win & Tie & Loss & $\Delta$ & Win & Tie & Loss & $\Delta$ & Win & Tie & Loss & $\Delta$ & \\
\midrule
MLLM-Judge        & 64.1 & 15.2 & 20.7 & +43.4 & 68.5 & 13.0 & 18.5 & +50.0 & 70.7 & 13.0 & 16.3 & +54.4 & 68.5 & 13.0 & 18.5 & +50.0 & +49.5 \\
Self-Refine       & 56.3 & 21.2 & 22.5 & +33.8 & 53.7 & 21.1 & 25.2 & +28.5 & 61.8 & 23.6 & 14.6 & +47.2 & 52.2 & 19.8 & 28.0 & +24.2 & +33.4 \\
Multi-Agent Debate & 58.9 & 12.0 & 29.1 & +29.8 & 55.7 & 13.0 & 31.3 & +24.4 & 57.8 & 14.1 & 28.0 & +29.8 & 55.7 & 13.0 & 31.3 & +24.4 & +27.1 \\
VISTA             & 55.1 & 16.8 & 28.1 & +27.0 & 55.1 & 14.5 & 30.4 & +24.7 & 64.0 & 15.7 & 20.3 & +43.7 & 57.3 & 14.6 & 28.1 & +29.2 & +31.2 \\
VQQA              & 50.7 & 23.9 & 25.4 & +25.3 & 52.8 & 25.0 & 22.2 & +30.6 & 51.7 & 22.8 & 25.4 & +26.3 & 45.2 & 26.1 & 28.7 & +16.5 & +24.7 \\
\midrule
Avg. $\Delta$             & \cellcolor{gray!30} & \cellcolor{gray!30} & \cellcolor{gray!30} & +31.9 & \cellcolor{gray!30} & \cellcolor{gray!30} & \cellcolor{gray!30} & +31.6 & \cellcolor{gray!30} & \cellcolor{gray!30} & \cellcolor{gray!30} & +40.3 & \cellcolor{gray!30} & \cellcolor{gray!30} & \cellcolor{gray!30} & +28.9 & +33.2 \\
\bottomrule
\end{tabular}
}}

\vspace{0.5em}

\subfloat[Pairwise comparison results on \dataset{} (Ours).]{
\resizebox{\textwidth}{!}{
\begin{tabular}{l|ccc c|ccc c|ccc c|ccc c|c}
\toprule
\multirow{2}{*}{\textbf{Method}} & \multicolumn{4}{c|}{\textbf{Accuracy}} & \multicolumn{4}{c|}{\textbf{Groundedness}} & \multicolumn{4}{c|}{\textbf{Completeness}} & \multicolumn{4}{c|}{\textbf{Helpfulness}} & \multirow{2}{*}{\textbf{Avg. $\Delta$}} \\
& Win & Tie & Loss & $\Delta$ & Win & Tie & Loss & $\Delta$ & Win & Tie & Loss & $\Delta$ & Win & Tie & Loss & $\Delta$ & \\
\midrule
MLLM-Judge        & 64.9 & 17.4 & 17.8 & +47.1 & 76.9 & 12.0 & 11.2 & +65.7 & 89.3 & 7.4 & 3.3 & +86.0 & 78.1 & 10.3 & 11.6 & +66.5 & +66.3 \\
Self-Refine       & 60.9 & 18.5 & 20.7 & +40.2 & 58.7 & 17.4 & 23.9 & +34.8 & 54.4 & 20.7 & 25.0 & +29.4 & 58.7 & 18.5 & 22.8 & +35.9 & +35.1 \\
Multi-Agent Debate & 54.5 & 26.0 & 19.4 & +35.1 & 59.1 & 22.7 & 18.2 & +40.9 & 70.2 & 14.0 & 15.7 & +54.5 & 62.0 & 20.7 & 17.4 & +44.6 & +43.8 \\
VISTA             & 54.6 & 22.7 & 22.7 & +31.9 & 60.7 & 18.0 & 21.3 & +39.4 & 63.6 & 19.7 & 16.7 & +46.9 & 59.0 & 19.2 & 21.8 & +37.2 & +38.9 \\
VQQA              & 65.3 & 16.0 & 18.7 & +46.6 & 68.0 & 17.3 & 14.7 & +53.3 & 80.1 & 11.8 & 8.1 & +72.0 & 66.5 & 17.3 & 16.2 & +50.3 & +55.6 \\
\midrule
Avg. $\Delta$             & \cellcolor{gray!30} & \cellcolor{gray!30} & \cellcolor{gray!30} & +40.2 & \cellcolor{gray!30} & \cellcolor{gray!30} & \cellcolor{gray!30} & +46.8 & \cellcolor{gray!30} & \cellcolor{gray!30} & \cellcolor{gray!30} & +57.8 & \cellcolor{gray!30} & \cellcolor{gray!30} & \cellcolor{gray!30} & +46.9 & +47.9 \\
\bottomrule
\end{tabular}
}}
\caption{\textbf{Text-Video Alignment} results. Win/Tie/Loss/$\Delta$ = Win - Loss (\%) of our method vs.\ baselines.}
\label{tab:pairwise}
\end{table*}

\begin{table*}[t!]
\centering
\subfloat[Pairwise comparison results on Vbench 2.0 dataset \citep{zheng2025vbench}.]{
\resizebox{\textwidth}{!}{
\begin{tabular}{l|ccc c|ccc c|ccc c|ccc c|c}
\toprule
\multirow{2}{*}{\textbf{Method}} & \multicolumn{4}{c|}{\textbf{Accuracy}} & \multicolumn{4}{c|}{\textbf{Groundedness}} & \multicolumn{4}{c|}{\textbf{Completeness}} & \multicolumn{4}{c|}{\textbf{Helpfulness}} & \multirow{2}{*}{\textbf{Avg. $\Delta$}} \\
& Win & Tie & Loss & $\Delta$ & Win & Tie & Loss & $\Delta$ & Win & Tie & Loss & $\Delta$ & Win & Tie & Loss & $\Delta$ & \\
\midrule
MLLM-Judge & 81.6 & 15.0 & 3.4 & +78.2 & 96.6 & 3.4 & 0.0 & +96.6 & 86.5 & 13.5 & 0.0 & +86.5 & 87.8 & 12.2 & 0.0 & +87.8 & +87.3 \\
Self-Refine  & 45.7 & 18.8 & 35.5 & +10.2 & 61.5 & 9.9 & 28.6 & +32.9 & 48.4 & 19.0 & 32.6 & +15.8 & 48.7 & 17.7 & 33.6 & +15.1 & +18.5 \\
Multi-Agent Debate & 50.8 & 21.7 & 27.5 & +23.3 & 65.6 & 16.0 & 18.4 & +47.2 & 53.8 & 19.3 & 26.9 & +26.9 & 55.7 & 19.7 & 24.6 & +31.1 & +32.1 \\
VISTA             & 35.9 & 31.9 & 32.2 & +3.7 & 52.3 & 18.2 & 29.5 & +22.8 & 40.6 & 30.5 & 28.9 & +11.7 & 37.2 & 29.2 & 33.6 & +3.6 & +10.5 \\
\midrule
Avg. $\Delta$     & \cellcolor{gray!30} & \cellcolor{gray!30} & \cellcolor{gray!30} & +28.9 & \cellcolor{gray!30} & \cellcolor{gray!30} & \cellcolor{gray!30} & +49.9 & \cellcolor{gray!30} & \cellcolor{gray!30} & \cellcolor{gray!30} & +35.2 & \cellcolor{gray!30} & \cellcolor{gray!30} & \cellcolor{gray!30} & +34.4 & +37.1 \\
\bottomrule
\end{tabular}
}}

\vspace{0.5em}

\subfloat[Pairwise comparison results on \dataset{}.]{
\resizebox{\textwidth}{!}{
\begin{tabular}{l|ccc c|ccc c|ccc c|ccc c|c}
\toprule
\multirow{2}{*}{\textbf{Method}} & \multicolumn{4}{c|}{\textbf{Accuracy}} & \multicolumn{4}{c|}{\textbf{Groundedness}} & \multicolumn{4}{c|}{\textbf{Completeness}} & \multicolumn{4}{c|}{\textbf{Helpfulness}} & \multirow{2}{*}{\textbf{Avg. $\Delta$}} \\
& Win & Tie & Loss & $\Delta$ & Win & Tie & Loss & $\Delta$ & Win & Tie & Loss & $\Delta$ & Win & Tie & Loss & $\Delta$ & \\
\midrule
MLLM-Judge        & 73.1 & 11.8 & 15.1 & +58.0 & 88.2 & 5.3 & 6.5 & +81.7 & 76.3 & 8.6 & 15.1 & +61.2 & 76.3 & 8.6 & 15.1 & +61.2 & +65.5 \\
Self-Refine       & 48.4 & 19.3 & 32.3 & +16.1 & 46.2 & 12.9 & 40.9 & +5.3 & 41.9 & 19.4 & 38.7 & +3.2 & 44.1 & 17.2 & 38.7 & +5.4 & +7.5 \\
Multi-Agent Debate & 63.4 & 12.9 & 23.7 & +39.7 & 66.7 & 10.7 & 22.6 & +44.1 & 66.7 & 10.7 & 22.6 & +44.1 & 67.7 & 11.9 & 20.4 & +47.3 & +43.8 \\
VISTA             & 47.8 & 28.9 & 23.3 & +24.5 & 61.1 & 23.3 & 15.6 & +45.5 & 47.8 & 20.0 & 32.2 & +15.6 & 48.9 & 27.8 & 23.3 & +25.6 & +27.8 \\
\midrule
Avg. $\Delta$     & \cellcolor{gray!30} & \cellcolor{gray!30} & \cellcolor{gray!30} & +34.6 & \cellcolor{gray!30} & \cellcolor{gray!30} & \cellcolor{gray!30} & +44.2 & \cellcolor{gray!30} & \cellcolor{gray!30} & \cellcolor{gray!30} & +31.0 & \cellcolor{gray!30} & \cellcolor{gray!30} & \cellcolor{gray!30} & +34.9 & +36.2 \\
\bottomrule
\end{tabular}
}}
\caption{\textbf{Temporal Consistency} results. Win/Tie/Loss/$\Delta$ = Win - Loss (\%) of our method vs.\ baselines.}
\label{tab:pairwise_consistency}
\end{table*}

\begin{table*}[t!]
\centering
\subfloat[Pairwise comparison results on Vbench 2.0 dataset \citep{zheng2025vbench}.]{
\resizebox{\textwidth}{!}{
\begin{tabular}{l|ccc c|ccc c|ccc c|ccc c|c}
\toprule
\multirow{2}{*}{\textbf{Method}} & \multicolumn{4}{c|}{\textbf{Accuracy}} & \multicolumn{4}{c|}{\textbf{Groundedness}} & \multicolumn{4}{c|}{\textbf{Completeness}} & \multicolumn{4}{c|}{\textbf{Helpfulness}} & \multirow{2}{*}{\textbf{Avg. $\Delta$}} \\
& Win & Tie & Loss & $\Delta$ & Win & Tie & Loss & $\Delta$ & Win & Tie & Loss & $\Delta$ & Win & Tie & Loss & $\Delta$ & \\
\midrule
MLLM-Judge        & 44.1 & 21.5 & 34.4 & +9.7 & 46.8 & 18.6 & 34.6 & +12.2 & 43.4 & 18.4 & 38.2 & +5.2 & 42.8 & 18.7 & 38.5 & +4.3 & +7.9 \\
Self-Refine       & 47.4 & 24.6 & 28.0 & +19.4 & 47.2 & 19.4 & 33.4 & +13.8 & 45.7 & 18.7 & 35.6 & +10.1 & 43.9 & 19.9 & 36.2 & +7.7 & +12.8 \\
Multi-Agent Debate & 44.8 & 23.0 & 32.2 & +12.6 & 50.9 & 18.7 & 30.4 & +20.5 & 48.8 & 18.7 & 32.5 & +16.3 & 48.8 & 19.6 & 31.6 & +17.2 & +16.7 \\
VISTA            & 42.9 & 21.3 & 35.8 & +7.1 & 40.3 & 20.0 & 39.7 & +0.6 & 43.5 & 19.7 & 36.8 & +6.7 & 42.3 & 20.3 & 37.4 & +4.9 & +4.8 \\
\midrule
Avg. $\Delta$     & \cellcolor{gray!30} & \cellcolor{gray!30} & \cellcolor{gray!30} & +12.2 & \cellcolor{gray!30} & \cellcolor{gray!30} & \cellcolor{gray!30} & +11.8 & \cellcolor{gray!30} & \cellcolor{gray!30} & \cellcolor{gray!30} & +9.6 & \cellcolor{gray!30} & \cellcolor{gray!30} & \cellcolor{gray!30} & +8.5 & +10.5 \\
\bottomrule
\end{tabular}
}}

\vspace{0.5em}

\subfloat[Pairwise comparison results on \dataset{}.]{
\resizebox{\textwidth}{!}{
\begin{tabular}{l|ccc c|ccc c|ccc c|ccc c|c}
\toprule
\multirow{2}{*}{\textbf{Method}} & \multicolumn{4}{c|}{\textbf{Accuracy}} & \multicolumn{4}{c|}{\textbf{Groundedness}} & \multicolumn{4}{c|}{\textbf{Completeness}} & \multicolumn{4}{c|}{\textbf{Helpfulness}} & \multirow{2}{*}{\textbf{Avg. $\Delta$}} \\
& Win & Tie & Loss & $\Delta$ & Win & Tie & Loss & $\Delta$ & Win & Tie & Loss & $\Delta$ & Win & Tie & Loss & $\Delta$ & \\
\midrule
MLLM-Judge        & 68.8 & 19.4 & 11.8 & +57.0 & 80.6 & 10.8 & 8.6 & +72.0 & 76.3 & 14.0 & 9.7 & +66.6 & 75.3 & 12.9 & 11.8 & +63.5 & +64.8 \\
Self-Refine       & 54.8 & 22.6 & 22.6 & +32.2 & 63.4 & 18.3 & 18.3 & +45.1 & 58.1 & 21.5 & 20.4 & +37.7 & 59.1 & 21.5 & 19.4 & +39.7 & +38.7 \\
Multi-Agent Debate & 67.7 & 19.4 & 12.9 & +54.8 & 73.1 & 17.2 & 9.7 & +63.4 & 71.0 & 16.1 & 12.9 & +58.1 & 71.0 & 18.2 & 10.8 & +60.2 & +59.1 \\
VISTA             & 51.1 & 32.2 & 16.7 & +34.4 & 56.7 & 27.7 & 15.6 & +41.1 & 52.2 & 33.4 & 14.4 & +37.8 & 52.2 & 33.4 & 14.4 & +37.8 & +37.8 \\
\midrule
Avg. $\Delta$     & \cellcolor{gray!30} & \cellcolor{gray!30} & \cellcolor{gray!30} & +44.6 & \cellcolor{gray!30} & \cellcolor{gray!30} & \cellcolor{gray!30} & +55.4 & \cellcolor{gray!30} & \cellcolor{gray!30} & \cellcolor{gray!30} & +50.1 & \cellcolor{gray!30} & \cellcolor{gray!30} & \cellcolor{gray!30} & +50.3 & +50.1 \\
\bottomrule
\end{tabular}
}}
\caption{\textbf{Physical Commonsense} results. Win/Tie/Loss/$\Delta$ = Win - Loss (\%) of our method vs.\ baselines.}
\label{tab:pairwise_commonsense}
\end{table*}

\begin{table*}[t!]
\centering
\resizebox{\textwidth}{!}{
\begin{tabular}{l|ccc c|ccc c|ccc c|ccc c|c}
\toprule
\multirow{2}{*}{\textbf{Method}} & \multicolumn{4}{c|}{\textbf{Accuracy}} & \multicolumn{4}{c|}{\textbf{Groundedness}} & \multicolumn{4}{c|}{\textbf{Completeness}} & \multicolumn{4}{c|}{\textbf{Helpfulness}} & \multirow{2}{*}{\textbf{Avg. $\Delta$}} \\
& Win & Tie & Loss & $\Delta$ & Win & Tie & Loss & $\Delta$ & Win & Tie & Loss & $\Delta$ & Win & Tie & Loss & $\Delta$ & \\
\midrule
MLLM-Judge        & 60.5 & 21.2 & 18.3 & +42.2 & 70.6 & 15.4 & 14.0 & +56.6 & 70.6 & 14.3 & 15.1 & +55.5 & 59.8 & 25.1 & 15.1 & +44.7 & +49.8 \\
Self-Refine       & 54.8 & 23.7 & 21.5 & +33.3 & 63.4 & 20.5 & 16.1 & +47.3 & 61.3 & 19.3 & 19.4 & +41.9 & 54.8 & 24.8 & 20.4 & +34.4 & +39.2 \\
Multi-Agent Debate & 53.8 & 23.6 & 22.6 & +31.2 & 65.6 & 19.3 & 15.1 & +50.5 & 67.7 & 17.2 & 15.1 & +52.6 & 55.9 & 24.7 & 19.4 & +36.5 & +42.7 \\
VISTA             & 48.5 & 20.2 & 31.3 & +17.2 & 55.6 & 15.6 & 28.8 & +26.8 & 60.8 & 18.2 & 21.0 & +39.8 & 50.5 & 30.1 & 19.4 & +31.1 & +28.7 \\
\midrule
Avg. $\Delta$     & \cellcolor{gray!30} & \cellcolor{gray!30} & \cellcolor{gray!30} & +31.0 & \cellcolor{gray!30} & \cellcolor{gray!30} & \cellcolor{gray!30} & +45.3 & \cellcolor{gray!30} & \cellcolor{gray!30} & \cellcolor{gray!30} & +47.5 & \cellcolor{gray!30} & \cellcolor{gray!30} & \cellcolor{gray!30} & +36.7 & +40.1 \\
\bottomrule
\end{tabular}
}
\caption{\textbf{Video-Audio Synchronization} results. Win/Tie/Loss/$\Delta$ = Win - Loss (\%) of our method vs.\ baselines on \dataset{}.}
\label{tab:pairwise_audio}
\end{table*}

\section{Implementation Details}

\subsection{Details of the Tools and Graph Construction}\label{sec:graph_construction}

Given the ordered keyframes $I_{1:K}$ and entity-state vocabulary $\mathcal{V}$, we construct one scene graph $G_k$ for each keyframe $I_k$. Each graph contains:
\begin{equation} G_k = \left( O_k,\, S_k,\, A_k,\, R_k^{\mathrm{sp}},\, R_k^{\mathrm{dep}},\, R_k^{\mathrm{act}} \right), 
\end{equation} 
where $O_k$ denotes the detected entities, $S_k$ their state assignments, $A_k$ the scene-level attributes, and $R_k^{\mathrm{sp}}$, $R_k^{\mathrm{dep}}$, and $R_k^{\mathrm{act}}$ the spatial, depth, and action relations, respectively. Processing keyframes in temporal order produces the graph sequence $\mathcal{G}=\{G_k\}_{k=1}^{K}$.

\paragraph{Entity States and Actions.} For each entity, the MLLM receives $I_k$ together with its state schema from $\mathcal{V}$ and assigns a value to each applicable state dimension. The output is restricted to the admissible values defined by the schema. Similarly, action relations are represented as subject--verb--object triples, where the verb must belong to the action vocabulary in $\mathcal{V}$. This shared vocabulary prevents synonymous descriptions across keyframes from being treated as different states or actions and enables symbolic comparison of their temporal transitions.

\paragraph{Tool Space $\Omega$ .} 
For visual tools, by default, entity states, action relations, and cardinalities for each $G_k$ are inferred directly by \textbf{Gemini 3 Flash} \citep{googledeepmind2025gemini3flash} from the keyframe, constrained to the vocabulary $\mathcal{V}$. Specialized perceptual tools are invoked only when a proposition is routed to them. Specifically, for depth-routed propositions, we use \textbf{SAM~3}~\citep{carion2026sam}, conditioned on the entity names, to obtain instance masks, bounding boxes, detection confidence, centers, and mask areas. We then apply \textbf{Depth Anything V2}~\citep{yang2024depth} to produce a depth map and compute pairwise relations such as \textit{in\_front\_of} and \textit{behind} over the localized entities. SAM~3 detections may also support explicit symbolic count queries when present, although cardinality is handled by the MLLM in the default path. For shot-scale propositions, we use a \textbf{DINOv2-based shot-scale classifier}~\citep{aslakey_shot_scale}. 

For audio tools, we use \textbf{NVIDIA's Sortformer} ({nvidia/diar\_sortformer\_4spk-v1}) \citep{park2024sortformer} for speaker diarization, producing timestamped speech segments and anonymous speaker labels for up to four speakers. For other sounds, including object sounds, music, and environmental noise, we use \textbf{Gemini} to extract foreground audio events with their start time, end time, description, and confidence. All detected events retain their original timestamps for alignment with the visual timeline.

For motion propositions, we experimented with several off-the-shelf motion estimators. However, these tools primarily capture low-level pixel displacement and generalize poorly to open-ended semantic motion requirements, often producing false negatives under camera movement, occlusion, or motion blur. We therefore route motion propositions to \textbf{Gemini 3}, which directly evaluates the corresponding video segments for the queried object or camera motion.

The resulting relation set is:
\begin{equation} 
R_k = R_k^{\mathrm{sp}} \cup R_k^{\mathrm{dep}} \cup R_k^{\mathrm{act}}. 
\end{equation} 

\emph{It is worth noting that, although we instantiate $\Omega$ with strong current models, \model{} is tool-agnostic: each verifier can be replaced by a stronger specialist or future omni-modal model, provided that it returns temporally localized confidence evidence through the same interface.}

\paragraph{Conditional Construction.} To avoid invoking unnecessary perceptual tools, graph construction follows the routes required by the compiled propositions. If segmentation-based evidence is needed, such as for counting or depth ordering, we construct the full tool-augmented graph described above. Otherwise, we use a lightweight path that instantiates the prompt entities and populates only their MLLM-inferred states and action relations, leaving unused object, spatial, depth, and scene-attribute fields empty. Thus, every $G_k$ exposes a common interface while containing only the evidence required for its propositions.

\subsection{Details of the Unified Specification Language} \label{app:unified_language}
As introduced in the main text, each proposition $\phi_i=(p_i,\tau_i,\omega_i)$ is governed by parameterized temporal operators and specific evidence routes. The temporal operator $\tau_i$ is drawn from Linear Temporal Logic (LTL). The operators \textsc{Always} and \textsc{Eventually} may be restricted to an absolute temporal window $[t_{\mathrm{start}},t_{\mathrm{end}}]$. The \textsc{Until} operator references a trigger proposition and may strictly bound the response delay or the number of permitted occurrences. Unparameterized operators default to evaluating over the full video duration.The route $\omega_i$ determines both the evidence modality and the specific tool drawn from $\Omega$. Predicates are grounded differently based on their modality: visual predicates are evaluated on static keyframes or dynamic video segments; auditory predicates are evaluated on timestamped sound events mapped to the nearest keyframes; and cross-modal predicates are evaluated by comparing each sound event against its concurrent visual segment.



\subsection{Details of Our \dataset{} Construction and Annotation} \label{subsec:dataset-annotations}

\paragraph{Prompt Creations.} 
We synthesize prompts using a SOTA LLM~\citep{googledeepmind2026gemini31pro} under a constrained generation protocol inspired by the logical depth of FOLIO~\citep{han2024folio}. We first define the prompt style: each prompt must contain at least one explicit real-time constraint, expressed either as a response deadline (e.g., ``within $N$ seconds of $X$'') or an absolute temporal window; at least three interdependent temporal or causal requirements; at least one underdetermined case whose trigger may or may not occur; and at least one audio or cross-modal requirement. All requirements must be visually concrete and plausible. We then prompt the LLM to generate prompts following five interacting-logic families: \emph{interacting until}, \emph{conditional exception}, \emph{conjunctive interlock}, \emph{scoped invariant chain}, and \emph{quantified interlock}. These families instantiate established temporal-property patterns, including scoped invariance, response and precedence, persistence, Boolean composition, and quantified requirements~\citep{dwyer1998property,manna1990hierarchy}. We vary prompt difficulty from simpler triggered-persistence and cascading-response patterns to more complex specifications containing between three and ten interacting clauses. Finally, because these initial prompts often resemble formal specifications rather than natural user requests, we use the same LLM to rewrite them in more natural, human-like language while preserving all requirements. We then employ human annotators then review the rewritten prompts and make any necessary corrections. After generating and removing bad prompts, we finally obtain 200 prompts balanced across modalities and difficulty levels.

\paragraph{Video Generation.} 
For each prompt, we generate one video using \textbf{Veo 3.1}~\citep{deepmind2025veo3} and one using \textbf{Seedance 2.0}~\citep{seedance2}, yielding 400 videos in total.

\paragraph{Video Annotations.}
Three highly qualified annotators with experience in NLP and CV research participate in our annotation process. For each video, two annotators jointly assess text--video alignment, physical commonsense, temporal consistency, and video--audio synchronization on a continuous scale from $0.0$ to $1.0$, following dimension-specific scoring definitions and quality intervals. Scores in \([0.0,0.2)\) indicate that the video largely fails the dimension’s requirements, \([0.2,0.4)\) that it meets some requirements but exhibits major violations, \([0.4,0.6)\) that it meets several requirements but has clear shortcomings, \([0.6,0.8)\) that it meets the main requirements with minor errors, and \([0.8,1.0]\) that it meets the requirements with at most negligible errors. The annotators are encouraged to consider multiple aspects of each dimension and weigh the video’s strengths alongside any errors they observe but not overly focusing on the errors. See \Cref{fig:qualitative_analysis} for examples. The two annotators must reach consensus before recording each score. A third annotator subsequently reviews all scores and discusses any questionable cases with the initial pair to determine the final annotations.

\section{MLLM Judge Prompts} \label{appdx:mllm-prompts}

\subsection{Alignment}
\begin{Verbatim}
You are an expert auditor comparing two automated video-evaluation systems on the quality of their TEXT-VIDEO ALIGNMENT feedback.

The ONLY task these evaluators should be doing is: checking whether the VIDEO faithfully depicts what the PROMPT describes. This means verifying presence of described objects, actions, spatial arrangements, temporal ordering, and other explicit prompt requirements.

IMPORTANT: Comments about general video quality (artifacts, flickering, deformation, smoothness, resolution, animation quality, cinematic quality) are IRRELEVANT to text-video alignment unless they directly prevent a prompt requirement from being recognizable. Do NOT reward an evaluator for pointing out visual defects that have nothing to do with whether the prompt's content is depicted. Penalize evaluators that conflate video quality with prompt alignment.

Video prompt: "{prompt}"

Watch the video. Then read the alignment feedback produced by two evaluators. For each criterion, decide which evaluator's feedback is better STRICTLY in terms of text-video alignment assessment.

Evaluator A:
{feedback_a}

Evaluator B:
{feedback_b}

For EACH criterion, choose which evaluator is better ("A" or "B"):

1. ACCURACY — Whose claims about whether specific prompt requirements are met or missing are more factually correct? Penalize hallucinated issues, incorrect verdicts, and irrelevant video-quality complaints disguised as alignment issues.
2. GROUNDEDNESS — Whose claims are more specifically tied to verifiable prompt requirements (named objects, described actions, stated relationships) rather than generic quality observations?
3. COMPLETENESS — Whose feedback covers more of the prompt's actual requirements (objects, actions, scene elements, temporal ordering)?
4. HELPFULNESS — Whose feedback more clearly explains which prompt requirements are satisfied and which are not, with specific evidence?

Do NOT reward feedback merely for being longer. Ignore any feedback content that is not about text-video alignment (e.g., comments on rendering quality, visual artifacts, animation smoothness, cinematic quality). Only judge based on the portions of feedback that assess whether prompt requirements are fulfilled.

Return JSON only:
```json
{{
  "accuracy": {{"winner": "A" or "B", "reason": "<reasoning of why>"}},
  "groundedness": {{"winner": "A" or "B", "reason": "<reasoning of why>"}},
  "completeness": {{"winner": "A" or "B", "reason": "<reasoning of why>"}},
  "helpfulness": {{"winner": "A" or "B", "reason": "<reasoning of why>"}}
}}
```
\end{Verbatim}

\subsection{Consistency}
\begin{Verbatim}
You are an expert auditor comparing two automated video-evaluation systems on the quality of their TEMPORAL CONSISTENCY feedback.

The ONLY task these evaluators should be doing is: assessing whether entities, objects, and backgrounds maintain consistent appearance, identity, and attributes throughout the video. This includes checking for identity changes, color shifts, objects appearing/disappearing, and visual continuity across frames.

IMPORTANT: Comments about text-video alignment or general aesthetic quality are IRRELEVANT unless they directly relate to consistency violations.

Video prompt: "{prompt}"

Watch the video. Then read the consistency feedback produced by two evaluators.

Evaluator A:
{feedback_a}

Evaluator B:
{feedback_b}

For EACH criterion, choose which evaluator is better ("A" or "B"):

1. ACCURACY — Whose claims about consistency issues are more factually correct when watching the video? Penalize hallucinated issues or missed real inconsistencies.
2. GROUNDEDNESS — Whose claims reference specific visual elements, timestamps, or frame regions rather than vague statements?
3. COMPLETENESS — Whose feedback identifies more actual consistency issues present in the video?
4. HELPFULNESS — Whose feedback more clearly explains what is inconsistent, where, and how?

Do NOT reward feedback merely for being longer.

Return JSON only:
```json
{{
  "accuracy": {{"winner": "A" or "B", "reason": "<reasoning of why>"}},
  "groundedness": {{"winner": "A" or "B", "reason": "<reasoning of why>"}},
  "completeness": {{"winner": "A" or "B", "reason": "<reasoning of why>"}},
  "helpfulness": {{"winner": "A" or "B", "reason": "<reasoning of why>"}}
}}
```
\end{Verbatim}

\subsection{Physical Commonsense}
\begin{Verbatim}
You are an expert auditor comparing two automated video-evaluation systems on the quality of their COMMONSENSE & PHYSICS feedback.

The ONLY task these evaluators should be doing is: assessing whether the video obeys physical laws and commonsense expectations — gravity, object permanence, anatomical correctness, realistic interactions, plausible motion dynamics, and cause-effect relationships.

IMPORTANT: Comments about text-video alignment or aesthetic quality are IRRELEVANT unless they directly relate to physics/commonsense violations.

Video prompt: "{prompt}"

Watch the video. Then read the commonsense/physics feedback produced by two evaluators.

Evaluator A:
{feedback_a}

Evaluator B:
{feedback_b}

For EACH criterion, choose which evaluator is better ("A" or "B"):

1. ACCURACY — Whose claims about physics/commonsense violations are more factually correct? Penalize false claims of violations that don't exist, or missing obvious ones.
2. GROUNDEDNESS — Whose claims reference specific physical phenomena, objects, or interactions rather than generic statements?
3. COMPLETENESS — Whose feedback identifies more actual physics/commonsense issues in the video?
4. HELPFULNESS — Whose feedback more clearly explains what violates physics/commonsense, why it's wrong, and what should happen instead?

Do NOT reward feedback merely for being longer.

Return JSON only:
```json
{{
  "accuracy": {{"winner": "A" or "B", "reason": "<reasoning of why>"}},
  "groundedness": {{"winner": "A" or "B", "reason": "<reasoning of why>"}},
  "completeness": {{"winner": "A" or "B", "reason": "<reasoning of why>"}},
  "helpfulness": {{"winner": "A" or "B", "reason": "<reasoning of why>"}}
}}
```
\end{Verbatim}

\subsection{Audio}
\begin{Verbatim}
You are an expert auditor comparing two automated video-evaluation systems on the quality of their AUDIO-VIDEO SYNCHRONIZATION feedback.

The ONLY task these evaluators should be doing is: assessing whether audio elements (speech, sound effects, music) are properly synchronized with visual events, whether audio quality is appropriate, and whether the audio matches the scene content.

IMPORTANT: Comments about visual quality or text-video alignment are IRRELEVANT unless they directly relate to audio-video sync issues.

Video prompt: "{prompt}"

Watch the video. Then read the audio-video sync feedback produced by two evaluators.

Evaluator A:
{feedback_a}

Evaluator B:
{feedback_b}

For EACH criterion, choose which evaluator is better ("A" or "B"):

1. ACCURACY — Whose claims about audio-video sync issues are more factually correct?
2. GROUNDEDNESS — Whose claims reference specific audio events and their corresponding visual moments?
3. COMPLETENESS — Whose feedback covers more of the audio-video relationships in the video?
4. HELPFULNESS — Whose feedback more clearly explains sync issues with specific evidence?

Do NOT reward feedback merely for being longer.

Return JSON only:
```json
{{
  "accuracy": {{"winner": "A" or "B", "reason": "<reasoning of why>"}},
  "groundedness": {{"winner": "A" or "B", "reason": "<reasoning of why>"}},
  "completeness": {{"winner": "A" or "B", "reason": "<reasoning of why>"}},
  "helpfulness": {{"winner": "A" or "B", "reason": "<reasoning of why>"}}
}}
```
\end{Verbatim}

\section{Methodology Prompts} \label{appdx:prompts}

\subsection{Explicit Proposition Compilation}
\begin{Verbatim}
You are a formal logic compiler. Convert the video prompt into a structured formal specification.

Available proposition verification tools (choose one per proposition):
- depth_tool:     Verifies depth-ordering relations: in_front_of, behind. Use only when the proposition explicitly describes one object behind/in front of another.
- shot_tool:      Verifies shot scale/type from a classifier: close-up, medium, wide, extreme close-up, two-shot. Use ONLY for propositions about shot framing/scale.
- motion_tool:    Verifies entity or background motion/movement (e.g. "trees swaying", "car driving", "water flowing", "person walking"). Use for propositions about objects or background elements moving, not camera movement.
- audio_tool:     Verifies sound events that must be heard rather than seen (e.g. "applause", "alarm ringing", "doorbell", "knock", "spoken words"). Use ONLY for propositions about audio/sound.
- vlm:            Verifies anything else — existence, actions, object states, attributes, spatial relations, or any visual/audio check no other tool handles.

Output JSON with these fields:
1. "propositions": dict mapping IDs (p1, p2, ...) to objects with:
   - "text": a simple, visually-verifiable statement in a single frame.
   - "temporal": one of "ALWAYS", "EVENTUALLY", or "UNTIL".
     * "ALWAYS": must hold in every frame (persistent objects, layout, spatial arrangements).
     * "EVENTUALLY": must hold in at least one frame (momentary actions or transitions).
     * "UNTIL": used ONLY for ordered event pairs — set on the SECOND event in a sequence.
       Also add "until_after": "<pid of the first event>" to express "first event holds UNTIL second event occurs".
   - "tool": the verification tool to use for this proposition (from the tool space above).
   - "count": (optional) expected cardinality constraint for the subject entity. Format: {{"cmp": "<comparator>", "n": <integer>}} where cmp is one of "=", ">", ">=", "<=", "<". Add this whenever the prompt implies a specific count (e.g. "a cat" -> {{"cmp":"=","n":1}}, "two dogs" -> {{"cmp":"=","n":2}}, "several birds" -> {{"cmp":">=","n":2}}). Omit for propositions about actions, states, or scene attributes.
   - TIMING FIELDS (all optional — add ONLY when the prompt states explicit clock times or time bounds):
     * "start_s": <float> absolute onset time in seconds. For ALWAYS/EVENTUALLY, restricts the check to frames at or after this time.
     * "end_s": <float> absolute offset time in seconds. For ALWAYS/EVENTUALLY, restricts the check to frames at or before this time.
       Example: "holds the cup from 5s to 8s" -> temporal "ALWAYS", start_s 5.0, end_s 8.0 (the state must hold continuously throughout that window).
       Example: "jumps between 3s and 5s" -> temporal "EVENTUALLY", start_s 3.0, end_s 5.0 (the event must occur at least once in that window).
     * "within_s": <float> relative deadline in seconds. For UNTIL only — the second event must occur within this many seconds after the first event (its until_after trigger).
       Example: "within 2 seconds of the doorbell, she opens the door" -> the "opens door" proposition has temporal "UNTIL", until_after the doorbell pid, within_s 2.0.
     * "min_repeats": <int> how many times an UNTIL trigger->response pair must recur (default 1). Set to 2+ when the prompt says "each", "every", "repeatedly", "multiple times".
     * "modality": "visual" (default) or "audio". Set "audio" when the event is a SOUND (applause, alarm, beep, ring, knock, spoken words) that must be HEARD rather than seen.
2. "nouns": list of all concrete, visually detectable object/entity noun phrases (e.g. "man", "woman", "car", "tree"). Exclude role labels, proper names, and character identifiers (e.g. exclude "Character A", "Character B", "Alice", "Bob", "character").
3. "camera_motion": "static" if the prompt explicitly states the camera is fixed or at a fixed position (e.g. "camera at eye level", "fixed camera", "static shot"), otherwise "dynamic".

Rules:
- Each proposition must express ONE verifiable fact. Do not mix subjects or unrelated predicates.
- Group related attributes of the same subject into one proposition (e.g. a character's full outfit = one proposition).
- Keep propositions minimal: one per key object, key action, or key spatial/depth relation.
- Each proposition must maintain the exact language as the original prompt.
- Use UNTIL only when the prompt explicitly implies one event precedes and enables another.
- Add timing fields ONLY when the prompt states explicit clock times, durations, or time bounds ("from 5s to 8s", "within 2 seconds", "after 3 seconds", "every 5 seconds"). Otherwise omit them entirely.
- A duration state ("X from Ts to Ts") is temporal ALWAYS with start_s and end_s (the state must persist across the whole window).
- A bounded momentary event ("X happens between Ts and Ts") is temporal EVENTUALLY with start_s and end_s.
- A relative deadline ("B within Ns of A") is temporal UNTIL on B, until_after A, with within_s N.
- Assign the most specific tool available; use vlm only as a last resort.
- The "tool" field MUST be one of: depth_tool, shot_tool, audio_tool, motion_tool, vlm.
- Use depth_tool ONLY for depth-ordering claims (in front of, behind).
- Use shot_tool ONLY for shot scale/framing (close-up, medium shot, wide shot, extreme close-up).
- Use camera_tool ONLY for camera or scene motion (static, panning, zooming, tracking, slow-motion).
- Use motion_tool for entity/background movement propositions (swaying, walking, flowing, driving).
- When structural arrangements (stacking, touching, containment, formation) are described, use the vlm tool to include overall structural propositions detailing which objects form the structure.
- Do NOT infer events beyond what is explicitly stated.

Example:
Prompt: "A cat jumps onto a table and knocks over a glass"
{{"propositions": {{
  "p1": {{"text": "A cat is visible.",            "temporal": "ALWAYS",     "tool": "vlm", "count": {{"cmp": "=", "n": 1}} }},
  "p2": {{"text": "The cat is jumping.",          "temporal": "EVENTUALLY", "tool": "vlm"}},
  "p3": {{"text": "The cat is on the table.",     "temporal": "UNTIL",      "tool": "vlm", "until_after": "p2"}},
  "p4": {{"text": "The glass is knocked over.",   "temporal": "UNTIL",      "tool": "vlm", "until_after": "p3"}}
}},
"nouns": ["cat", "table", "glass"],
"camera_motion": "dynamic"}}

Example (with timing):
Prompt: "A person walks into a kitchen. The person holds a cup from 5s to 8s. A doorbell rings, and within 2 seconds the person walks away."
{{"propositions": {{
  "p1": {{"text": "A person is visible.",              "temporal": "ALWAYS",     "tool": "vlm", "count": {{"cmp": "=", "n": 1}} }},
  "p2": {{"text": "The person walks into a kitchen.",  "temporal": "EVENTUALLY", "tool": "motion_tool"}},
  "p3": {{"text": "The person holds a cup.",           "temporal": "ALWAYS",     "tool": "vlm", "start_s": 5.0, "end_s": 8.0}},
  "p4": {{"text": "A doorbell rings.",                 "temporal": "EVENTUALLY", "tool": "vlm", "modality": "audio"}},
  "p5": {{"text": "The person walks away.",            "temporal": "UNTIL",      "tool": "motion_tool", "until_after": "p4", "within_s": 2.0}}
}},
"nouns": ["person", "kitchen", "cup", "doorbell"],
"camera_motion": "dynamic"}}

Prompt: {prompt}

Return the output in JSON format: 
```json
...
```
\end{Verbatim}

\subsection{Entity-State Vocabulary Construction}
\begin{Verbatim}
Given this video prompt and its keyframes, determine what physical states to track for each entity and what action relations are possible between entities.

Prompt: "{prompt}"
Entities: {nouns}

Look at the actual frames. For each entity, identify:
1. ALL state dimensions that are visually observable and could change across frames.
2. The possible action relations between entities as verbs.

Return JSON:
```json
{{
  "entity_states": {{
    "<entity>": {{
      "<dimension>": ["<value1>", "<value2>", ...]
    }}
  }},
  "action_vocabulary": ["<verb1>", "<verb2>", ...]
}}
```

Rules for entity_states:
- Include all visually observable state dimensions, not just those mentioned in the prompt.
- Values must be visually distinguishable from a single video frame.
- Keep values discrete and concrete (not vague like "normal" or "other").

Rules for action_vocabulary:
- List ALL possible action/relation verbs observable between the entities in this video.
- Verbs MUST be in base (infinitive) form: "hold", "wear", "stand_on", "sit", "walk", "look_at". NOT "holding", "wearing", "standing_on".
- Include structural relations: "have", "contain", "cover", "support".
- Include interactions: "chase", "follow", "push", "pull", "throw", "catch".
- Only include verbs that are visually verifiable from the frames.
- Keep verbs specific and grounded — avoid vague verbs like "interact" or "relate".
\end{Verbatim}

\subsection{Entity-State and Action Extraction.}

\begin{Verbatim}
Given these detected objects in the image:
{objects}

And this fixed action vocabulary (use ONLY these verbs):
{action_vocabulary}

For each pair of objects that has a visible relationship, output a triple [subject, verb, object] where verb is from the vocabulary above.
Do NOT include spatial relations (left, right, above, below) — those are computed separately.
Do NOT invent verbs outside the vocabulary. If no verb from the vocabulary applies, skip that pair.
Return as a JSON list of triples: [subject, relation, object].
Return the output in JSON format: ```json...```
\end{Verbatim}

\subsection{Implicit Physical Proposition Compilation}
\begin{Verbatim}
You generate commonsense physics rules for auditing an AI-generated video, following
Manna & Pnueli's temporal hierarchy. You are given the prompt and an EVENT-TRANSITION
narrative (how each entity's state changes across frames). Reason about the TRANSITIONS,
not isolated states.

DEFINITIONS:
  INVARIANT (always P): a property that must hold in EVERY frame. This includes:
    - irreversibility: once a state is reached it cannot revert (broken -> whole is impossible),
    - mutual exclusion: two states that can never co-occur,
    - count stability: a counted set cannot change.
    NOT pose/color/texture/appearance persistence.
  CAUSALITY (precedence): a state transition into an effect requires its enabling cause
    to be present AT THAT TRANSITION (same or adjacent frame), with spatial contact/proximity
    where physics demands it. E.g. a sword can only become "broken" if it is in CONTACT with
    something at the moment it breaks — "they were fighting earlier" is NOT enough.
  LIVENESS (response): a physical PROCESS that has started must eventually reach its natural
    completion, derived from PHYSICS (not from the prompt's wording). E.g. a lit fire must
    eventually go out; a falling object must land; a thrown object must come down.

CRITICAL — ONE RULE PER PHYSICAL EVENT:
- First identify the DISTINCT physical events/transitions in the narrative.
- For EACH event, emit EXACTLY ONE rule, assigned to the single best-fitting family.
  Never describe the same event with two or three rules across families.
- Family assignment guide:
    * An irreversible state change caused by an action (break, spill, burn, scatter, consume)
      -> CAUSALITY (does the effect have its contact-cause at the transition?). This already
      implies it cannot un-happen, so do NOT also write an invariant for it.
    * A process that must reach completion (falling->land, lit->extinguish) -> LIVENESS.
    * A static constraint with NO transition (count stability, mutual exclusion) -> INVARIANT.
- A causality rule must demand the cause co-occur with the effect transition (contact/proximity),
  not merely exist somewhere earlier.
- Do NOT generate prompt-restatement liveness (that is text-video alignment, not physics).
- Do NOT generate attribute/recognition rules ("stays green", "is a panda").

EXAMPLE
Prompt: "The panda grips a thick bamboo stalk with its front paws, then snaps it cleanly with
a powerful bite before bringing the broken section to its mouth to strip away the outer layer."
Event narrative: {{"bamboo section": {{"state": "whole -> broken -> stripped"}}, "panda": {{"jaws": "biting -> chewing"}}}}
Two distinct events -> exactly two rules (one each, no overlap):
  - CAUSALITY: "The bamboo can only become broken if the panda's jaws/paws are in direct contact with it at the moment of breaking." (the break event; severity critical)
  - CAUSALITY: "The outer layer can only become stripped if the panda's mouth is in contact with the bamboo during stripping." (the strip event; severity major)
Do NOT additionally write an invariant "broken can't revert" or a liveness "biting must break it" — those re-describe the same two events.
Bad rules: "the bamboo stays green" (attribute), "the panda is a panda" (recognition), "the panda eats bamboo" (prompt restatement).

Video prompt: "{prompt}"
Entities: {nouns}
Entity-state vocabulary (entity -> dimension -> values):
{schema}
Event-transition narrative (per entity, how states change across frames):
{events}

Emit AT MOST {max_inv} invariants, {max_caus} causality, {max_live} liveness.
Fewer, higher-significance rules are better. Return [] for a family if nothing
genuinely distinct remains to say. Each rule MUST include a "text" field.

Return JSON:
```json
{{
  "invariant": [
    {{"id":"inv_N","entity":"<e>","text":"<sentence>",
    "severity":"critical|major|minor","reason":"<why significant>"}}
  ],
  "causality": [
    {{"id":"caus_N","entity":"<e>","text":"<sentence>",
    "cause":"<what must co-occur at the transition>","severity":"critical|major|minor",
    "reason":"<why>"}}
  ],
  "liveness": [
    {{"id":"live_N","entity":"<e>","text":"<sentence>",
    "severity":"critical|major|minor","reason":"<why>"}}
  ]
}}
```
\end{Verbatim}

\subsection{Consistency Grounding}

\paragraph{Entity Grounding.}
\begin{Verbatim}
You are auditing the temporal consistency of entities across video keyframes. Your job is
to FIND DEFECTS, not to confirm everything is fine.

Entities to evaluate: {entities}

The keyframes are shown in temporal order, labeled [frame_XXXXX.jpg]. The first keyframe is
the reference appearance for each entity.

For EACH entity, scan every keyframe and identify the frames where the entity exhibits a
temporal-consistency DEFECT relative to its reference appearance. A defect is any of:
- morphing / shape distortion (geometry warps, parts change size or proportion)
- melting / merging / dissolving of body parts or structure
- texture "swimming" or flickering (surface details drift independently of motion)
- identity drift (the entity gradually turns into a different thing/class)
- parts appearing or disappearing (extra limbs, missing parts, duplicated parts)
- catastrophic deformation (entity becomes unrecognizable)

Do NOT count natural changes: pose changes, expression changes, normal motion, gradual
lighting, or the entity simply moving in/out of frame. Those are NOT defects.

For each entity, give:
- "score": overall temporal consistency in [0,1]:
    1.0  = flawless, no defects in any frame
    0.7-0.9 = minor flicker / slight texture swimming in a few frames
    0.4-0.6 = noticeable morphing or local deformation in some frames
    0.2-0.4 = severe morphing / melting / parts appearing or disappearing
    0.0-0.2 = entity changes class/type or becomes unrecognizable
- "error_frames": the list of frame names where defects occur (empty if none)
- "defect": one short sentence describing the worst defect (empty if none)

Return JSON, one entry per entity:
```json
{{
  "<entity_name>": {{"score": <float>, "error_frames": ["frame_XXXXX.jpg", ...], "defect": "<what is visibly wrong>"}}
}}
```
\end{Verbatim}

\paragraph{Background Grounding.}
\begin{Verbatim}
You are judging temporal consistency of the background/environment across video keyframes.

The first frame (frame 0) is the reference. For each subsequent frame, rate background consistency using this scale:
- 0.0-0.2: Background is completely different scene, or has obvious screen-shot switch.
- 0.2-0.4: Background has severe issues: elements appear/disappear, major structural changes, heavy flickering.
- 0.4-0.6: Background has noticeable flaws: texture flickering, unstable optical properties (brightness/contrast/color jumps).
- 0.6-0.8: Background is consistent with only minor differences (gradual natural lighting changes are fine).
- 0.8-1.0: Background is highly stable and smooth. Reserve 1.0 for flawless.

Check for these specific error points:
- Do background elements suddenly appear or disappear?
- Is there flickering color or unstable brightness/contrast/saturation?
- Are there speckles, black patches, or distortion?

Keyframes are labeled [frame_XXXXX.jpg]. The first keyframe is the reference. Rate each subsequent frame.

Return JSON:
```json
{{"scores": {{"frame_XXXXX.jpg": <float>, ...}}}}
```
\end{Verbatim}

\subsection{Visual and Motion Grounding}
\begin{Verbatim}
You are verifying statements against a video. The video's keyframes are shown below, labeled by their frame names.

Statements:
{statements}

For each statement and each keyframe, rate whether the statement holds in that frame. Be STRICT — every detail in the statement must be visually confirmed:
- 0.0-0.2: The described element is completely absent, or present but so deformed/morphed/corrupted that it is unrecognizable.
- 0.2-0.4: The element is barely recognizable but has severe visual defects (melting, morphing, missing parts, wrong shape).
- 0.4-0.6: The element is present but key details from the statement are missing or wrong (e.g. "dappled sunlight" but lighting is flat; "running" but the subject is stationary).
- 0.6-0.8: The element is present and most details match, with only minor discrepancies.
- 0.8-1.0: Every detail in the statement is accurately depicted. Reserve 1.0 for flawless match of ALL specifics.

IMPORTANT:
- Check EVERY specific detail in the statement (adjectives, adverbs, spatial descriptions, lighting, motion). Missing details should lower the score proportionally.
- An entity with severe morphing/melting that makes it unrecognizable as what the statement describes should score below 0.4. However, if the entity is still clearly identifiable despite rendering artifacts (distortion, blur, slight deformation), score 0.6 or above — rendering quality is not the focus, content alignment is.
- Descriptive details (lighting quality, motion, texture) must be visually confirmed, not assumed.

Keyframes: {frame_names}

Output JSON only — scores per statement per frame, and for EVERY statement (whether passing or failing), a brief reason noting which specific details are satisfied and which are missing:
```json
{{
  "scores": {{
    "<pid>": [<score_frame_0>, <score_frame_1>, ..., <score_frame_N>]
  }},
  "reasons": {{
    "<pid>": "<which details are satisfied; which details are missing or wrong>"
  }}
}}
```
\end{Verbatim}

\subsection{Audio-Event Extraction}
\begin{Verbatim}
Listen to this audio track from a video. Identify occurrences of these specific sounds:
{vocab_str}

Also report any other distinct foreground sounds you hear that are NOT in the list above.

For each sound event, report:
- start: when it begins (seconds)
- end: when it ends (seconds)
- sound: use the EXACT name from the list above if it matches, otherwise describe freely
- confidence: how certain you are (0.0-1.0)

Return JSON only:
```json
{{
  "sounds": [
    {{"start": <float>, "end": <float>, "sound": "<description>", "confidence": <float>}}
  ]
}}
```

Rules:
- Only report sounds you actually hear.
- If the same sound spans a continuous uninterrupted duration, report it as one entry.
- Do NOT report ambient/background sounds (e.g., "quiet ambient noise", "room tone", "silence") — only report foreground sound events.
- Sort entries by start time.
\end{Verbatim}

\subsection{Video--Audio Sync Grounding}
\begin{Verbatim}
You are watching a SHORT VIDEO CLIP (no audio). A sound detection system reports that
the following sound occurs during this exact time window:

Sound: "{sound}"

Watch the visual content carefully and judge whether the visual is consistent with this sound
at this exact moment.

Return JSON only:
```json
{{
  "score": <float 0.0-1.0>,
  "reasoning": "<brief: what visual evidence supports or contradicts the sound>"
}}
```

Scoring guide:
- 0.0-0.2: The visual directly contradicts the sound — the scene makes this sound physically impossible (e.g., the sound source cannot exist in this environment).
- 0.2-0.4: No visual evidence of the sound source, and the scene context provides no plausible explanation for the sound.
- 0.4-0.6: The sound is contextually plausible (the scene could reasonably produce it) but the source is not directly visible in the clip, OR the sound source is visible but clearly not active at this moment.
- 0.6-0.8: The sound source is clearly present and active in the visual, with the action closely aligned to the sound's timestamp.
- 0.8-1.0: The visual precisely and unambiguously shows the sound source producing the sound at exactly this moment.
\end{Verbatim}

\subsection{Perception-Aware Calibration}
\begin{Verbatim}
{examples}

You are judging whether each proposition is PERCEPTUALLY SIGNIFICANT to a human viewer.

Prompt: {prompt}

For each proposition below, classify as:
  1 = SIGNIFICANT: A human viewer would notice and care if this is violated.
     (e.g., main subject missing, physics violation, wrong action, object deformed)
  0 = NOISE: A human viewer would NOT notice or care about this.
     (e.g., micro-flicker, background texture change, minor color shift, abstract state label)

Propositions:
{items}

Return a JSON array: [{{"id": "<id>", "label": 0 or 1}}, ...]

\end{Verbatim}

\subsection{Report Generation}

\paragraph{Report Generation.}
\begin{Verbatim}
You are a video evaluation review generator. You synthesize a final review report from structured verification findings. All findings are expressed as propositions with verdicts.

Video prompt: "{prompt}"

## Alignment Propositions
{alignment_log}

## Commonsense Propositions
{commonsense_log}

## Consistency Propositions
{consistency_log}

## Audio Propositions
{audio_log}

---

## Review Guidelines

The propositions above are the verified evidence. Base your report on them, following these principles:

**Grounding**
Ground every claim in the proposition verdicts and their timestamps. When you describe a failure, say what was violated, when, and include the specific reasoning about what is wrong (from the proposition's reason or video itself). 

**Completeness**
Cover the meaningful findings in each dimension — both what the video does well and where it fails. Do not omit a real violation, and do not invent one to seem thorough.

**Clarity**
Write a comprehensive, accurate, and grounded report for each dimension. Assume the reader has not seen the verification trace. Do not reference internal proposition IDs (e.g., p1, va_1, cons_bg) or numerical scores — describe the findings in natural language.

**Judgment**
Use judgment about the propositions themselves. If a proposition is incorrect, irrelevant to the video, or impossible to assess (e.g., lip-sync when there is no speech, or an attribute of an entity that does not appear), set it aside rather than forcing a verdict into the report. Similarly, if a finding does not make physical or logical sense (e.g., claiming "plumage" and "wingspan" are separate entities that independently lose consistency), recognize it as a measurement artifact and do not report it as a real issue. Do not reference internal proposition IDs (e.g., p1, va_1, cons_bg) — describe the findings in natural language.

## Output Format

Write a unified evaluation report with exactly 4 sections:

### 1. Text-Video Alignment
Assess how well the video realizes the prompt based on proposition verification results.

### 2. Temporal Consistency
Assess entity and background stability over time based on consistency findings.

### 3. Commonsense & Physics
Assess physical plausibility based on rule violations and visual/audio checks.

### 4. Audio-Video Synchronization
Assess audio-visual alignment based on audio findings.

For each section: state what works, what fails (with timestamps), and severity. Dismiss any propositions that do not make sense for this video. Be direct and factual.
\end{Verbatim}

\end{document}